\documentclass{article}

 \PassOptionsToPackage{numbers,sort&compress}{natbib}
 \usepackage[final,main]{neurips_2026}

\usepackage[utf8]{inputenc} 
\usepackage[T1]{fontenc}    
\usepackage{hyperref}       
\usepackage{url}            
\usepackage{booktabs}       
\usepackage{amsfonts}       
\usepackage{amsmath}        
\usepackage{amssymb}        
\usepackage{nicefrac}       
\usepackage{microtype}      
\usepackage{xcolor}         
\usepackage{graphicx}       
\usepackage{multirow}       
\usepackage{pifont}         
\usepackage{tabularx}       

\newcommand{\R}{\mathbb{R}}
\newcommand{\partk}[1]{\texttt{<part\_#1>}}

\title{3D-PLOT-LLM: Part-Level Object Tokens for 3D Large Language Models}

\author{%
  Jintang Xue \\
  University of Southern California \\
  Los Angeles, California, USA\\
  \texttt{jintangx@usc.edu} \\
  \And
  Xinyu Wang \\
  University of Southern California \\
  Los Angeles, California, USA\\
  \texttt{xwang350@usc.edu} \\
  \AND
  Yixing Wu \\
  University of Southern California \\
  Los Angeles, California, USA\\
  \texttt{yixingwu@usc.edu} \\
  \And
  Jingwen Chen \\
  Ohio State University \\
  Columbus, Ohio, USA \\
  \texttt{chen.15073@buckeyemail.osu.edu} \\
  \And
  C.-C. Jay Kuo \\
  University of Southern California \\
  Los Angeles, California, USA\\
  \texttt{jckuo@usc.edu} \\
}

\begin{document}

\maketitle

\begin{abstract}
3D multimodal large language models (3D MLLMs) describe a 3D
object as a whole but cannot address, name, or reason about
its parts. Prior part-aware attempts add segmentation
decoders, heavier 3D encoders, or bounding-box grammars at
substantial parameter cost. We take a fundamentally different
path: we reorganize the input token stream so that parts
become directly addressable through the LLM's own vocabulary.
Our model, \textbf{3D-PLOT-LLM}, partitions the frozen point
encoder's patches into $K$ locally coherent regions and
inserts, before each region's patch tokens, a learnable
per-region marker and a reserved vocabulary
token~\partk{k}; a Marker-Space Refinement (MSR) module then
conditions each marker on its region's spatial statistics
and adjacency neighbors. The model thus cites parts in its
output and follows prompts that refer to parts by token, a
capability absent from prior object-level 3D MLLMs. To probe
this interface, we construct \textbf{PartVerse-QA}, a
vocabulary-level part-QA benchmark adapted from PartVerse
mesh annotations ($77$K training pairs and $588$ held-out
queries on disjoint object splits), on which 3D-PLOT-LLM
reaches caption-to-slots Jaccard $0.459$ and
Exact-match $13.78\%$ ($+64\%$ relative over the strongest
non-MSR variant), with a slot-to-caption GPT-4o judge of
$44.68$. On the text-description track of the 3DCoMPaT-GrIn part-aware grounded
description benchmark, 3D-PLOT-LLM outperforms PointLLM,
Kestrel, PARIS3D, and SegPoint on every reported text metric,
and ShapeLLM on $3$ of $4$, with up to $+3.03$ GPT-4o judge
over PointLLM. On Objaverse whole-object captioning, adding
PartVerse-QA at Stage~2 yields $+0.65$ SBERT and $+1.85$
GPT-4o over PointLLM, and tops PointLLM-PiSA on $4$ of $5$
traditional metrics (SBERT, SimCSE, BLEU-1, METEOR) despite
targeting a different (part-grounded) objective.
All with under $1$M new trainable parameters on a frozen
point encoder, an order of magnitude below prior part-aware
3D MLLMs, and no learned part detector, segmentation decoder, or bounding-box head.
\end{abstract}

\section{Introduction}
\label{sec:intro}

\begin{figure}[!t]
  \centering
  \includegraphics[width=0.95\linewidth]{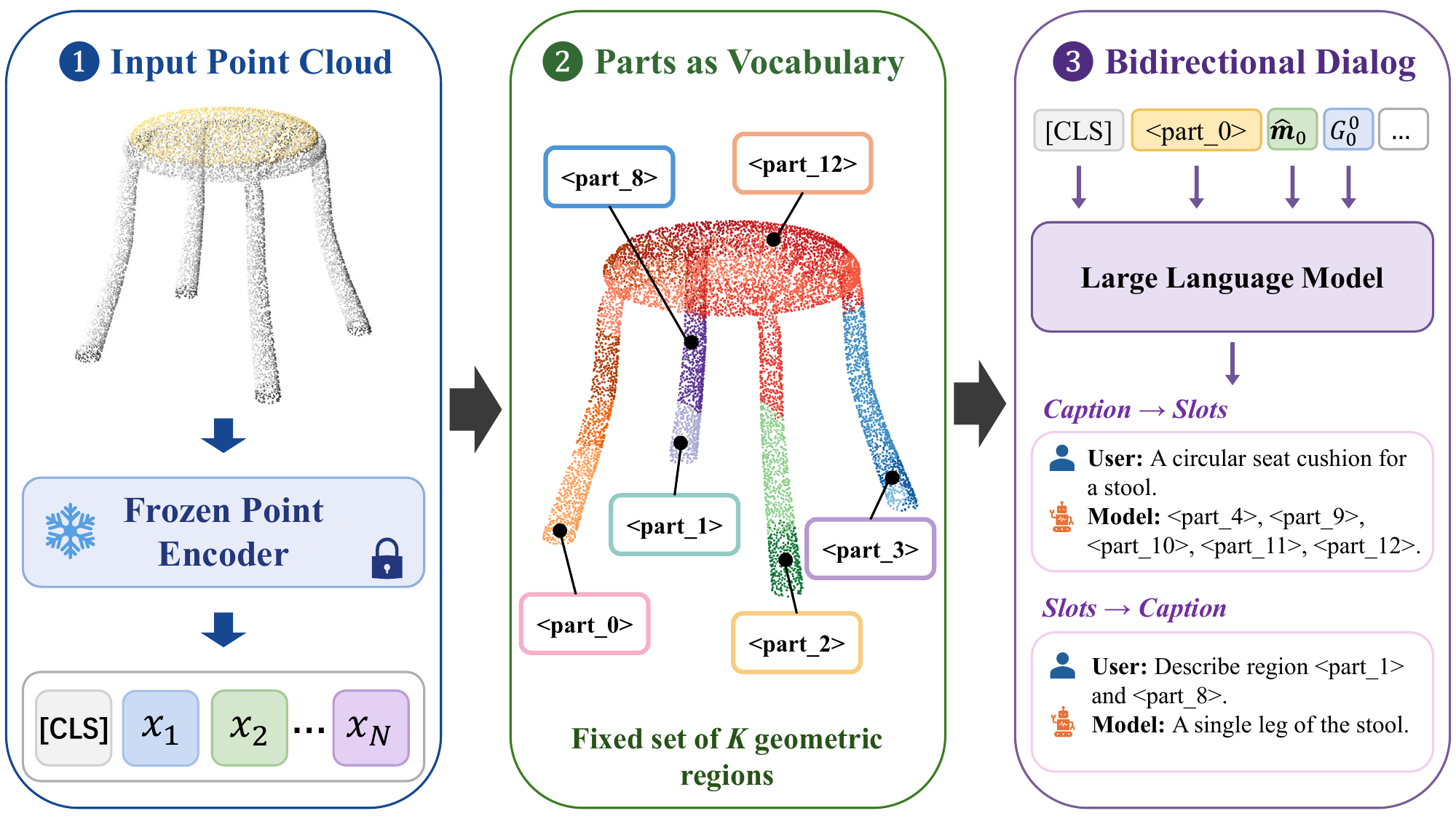}
  \caption{\textbf{3D-PLOT-LLM} treats each geometric region of a 3D
  object as a first-class addressable token in the LLM's
  vocabulary. A frozen point encoder's patch tokens are
  partitioned into a fixed set of $K$ geometric regions
  (center panel), and a reserved vocabulary slot~\partk{k}
  makes each region readable and writable through the same
  interface the LLM uses for text (right panel). No
  segmentation decoder is used; the interface addresses coarse geometric regions, not point-wise masks.}
  \label{fig:teaser}
\end{figure}

3D multimodal large language models (3D MLLMs) align point
clouds with language for free-form object-level understanding
(e.g., captioning, classification,
QA~\citep{xu2024pointllm,qi2024shapellm,guo2023point,guo2026pisa}), with a parallel
branch lifting MLLMs to whole-scene
grounding~\citep{hong20233d,huang2024chat,zemskova20253dgraphllm,mao2025spatiallm,chen2024grounded,thomas2025pts3d}.
At the object level, however, 3D MLLMs still present each
object as a flat sequence of patch tokens, with no
compositional units to refer to, name, or aggregate over at
the part level: PointLLM~\citep{xu2024pointllm} can caption a chair
but has no token denoting its back or legs. Prior part-aware
object 3D MLLMs pair the LLM with dense-prediction machinery
(Kestrel~\citep{ahmed2025kestrel}: a segmentation decoder driven by
LISA/GLaMM~\citep{lai2024lisa,rasheed2024glamm} \texttt{[SEG]} tokens;
Part-X-MLLM~\citep{wang2025part}: a heavier dual-pathway encoder
with a bounding-box grammar), inflating the non-LLM parameter
budget by roughly an order of magnitude. More fundamentally,
neither changes the LLM's input vocabulary, so parts remain
outputs of an external module rather than entries the LLM can
read, emit, and reason over in its own token stream; their dense predictions are strong, but that axis is orthogonal to the interface question we address. Our model, \textbf{3D-PLOT-LLM},
closes this gap by making parts addressable tokens in
the LLM's vocabulary (Figure~\ref{fig:teaser}).

A natural reflex would be to define a canonical part
decomposition and reserve a vocabulary slot per semantic part.
We argue against this on three grounds: (i) dense part
annotation does not scale, since existing human annotations use
category-specific schemas and reach at most tens of thousands
of objects, far below the $\sim\!660$K-object alignment regime
of 3D MLLMs~\citep{xu2024pointllm}; relying on other pretrained
modules just relocates this cost upstream. (ii) Part
granularity is intrinsically ambiguous, since a chair admits
many valid hierarchies (seat$+$back; $+$armrests; $+$four legs),
each reasonable. (iii) Captioning does not require a
semantic-category label; ``a wooden cylindrical pole at the
bottom'' is a useful answer whether the part is labeled ``stool
leg'' or ``support post''.

We instead position parts as emerging from input-side token
organization rather than architectural decomposition. We
partition each object's patch tokens into a fixed set of $K$
geometric regions
via a deterministic, training-free procedure that grows regions over patch spatial coordinates and
frozen point encoder features (App.~\ref{app:bcp}), and
reserve a per-object vocabulary
\partk{0},\dots,\partk{K{-}1} in the LLM's tokenizer. Part
identity emerges from data as a learned composition of
these region tokens; a nameable part (``the shovel handle'',
``the boat hull'') is expressed as a set of \partk{k} tokens,
learned end-to-end from PartVerse-QA. Each region also
carries a learnable marker $\mathbf{m}_k \in \R^{384}$, refined
by a lightweight Marker-Space Refinement (MSR) module
conditioning the marker on the region's spatial statistics and
adjacency. With under $1$M new trainable parameters on a
frozen $22$M-parameter point encoder, this part-level interface
(absent from prior object-level 3D MLLMs) yields consistent
gains across PartVerse-QA, the text-description metrics of 3DCoMPaT-GrIn, and Objaverse
captioning (\S\ref{sec:exp}). Our contributions:\footnote{Code, weights, and PartVerse-QA: \url{https://github.com/jintangxue/3D-PLOT-LLM}.}
\begin{itemize}
    \item \textbf{Part-aware addressing without canonical
    decomposition.} 3D-PLOT-LLM reserves per-object vocabulary
    tokens \partk{0}\dots\partk{K{-}1} tied to a
    deterministic, training-free spatially coherent partition of frozen-encoder patches; the LLM learns part identity
    as compositions of these tokens and uses them as a
    bidirectional interface it both reads and emits, without a learned part detector or segmentation decoder for pre-segmented objects.
    \item \textbf{Marker-Space Refinement (MSR).} A lightweight
    residual module conditioning per-region markers
    ($\R^{384}$) on region statistics and inter-region
    adjacency, supplying the structural context that binds
    \partk{k} embeddings to region content.
    \item \textbf{PartVerse-QA: a bidirectional caption-to-slots
    (C2S) and slot-to-caption (S2C) part-addressing benchmark.}
    Adapted from PartVerse~\citep{dong2025one} mesh annotations:
    $77$K training pairs $+$ $588$ held-out queries ($392$ C2S,
    $196$ S2C) on object-disjoint splits, aligned to Objaverse
    $8192$-point clouds and quality-filtered
    (App.~\ref{app:partverse}).
    \item \textbf{Empirical evaluation across three
    benchmarks.} PartVerse-QA;
    the part-aware point grounded description task (PaPGD) of 3DCoMPaT-GrIn~\citep{ahmed2025kestrel,slim20253dcompat++}, compared with PointLLM, ShapeLLM~\citep{qi2024shapellm},
    Kestrel~\citep{ahmed2025kestrel},
    PARIS3D~\citep{kareem2024paris3d}, and
    SegPoint~\citep{he2024segpoint};
    and Objaverse captioning, compared with PointLLM, ShapeLLM, ENEL~\citep{tang2026exploring}, and
    PiSA~\citep{guo2026pisa} (re-evaluated and cited rows are marked in the tables).
\end{itemize}

\section{Related work}
\label{sec:related}

\paragraph{Object-level 3D MLLMs.}
Object-level 3D MLLMs share a recipe: project patch tokens from
a point encoder into the LLM input space and train on
large-scale point-text supervision.
PointLLM~\citep{xu2024pointllm} exemplifies this with a
Point-BERT~\citep{yu2022point} encoder aligned to
Vicuna~\citep{chiang2023vicuna} via a two-stage protocol (660K
Cap3D~\citep{luo2023scalable} captions of Objaverse~\citep{deitke2023objaverse} for
alignment, then 70K GPT-generated complex instructions for
tuning); PiSA~\citep{guo2026pisa} extends it with a self-augmented
data engine and PiSA-Bench. ENEL~\citep{tang2026exploring} removes the point encoder altogether and trains on the same PointLLM corpus. Other systems vary specific
ingredients: GPT4Point~\citep{qi2024gpt4point} substitutes a Q-Former
projector and scales to $\sim$1M Objaverse-XL captions;
ShapeLLM~\citep{qi2024shapellm} swaps in a ReCon++ encoder for
embodied interaction; Point-Bind~\citep{guo2023point} aligns
points to ImageBind~\citep{girdhar2023imagebind} space. A parallel
foundation-encoder line, notably Uni3D~\citep{zhou2023uni3d}
(billion-parameter 2D-initialized ViT aligned with image-text
features), supplies backbones to several of these MLLMs.
Across the landscape, no object MLLM exposes an addressable
part vocabulary; each treats the object as a flat token stream
and cannot cite specific parts in its answer.

\paragraph{Part-aware object 3D MLLMs.}
Prior systems fall along two design lines. Mask-based
methods inherit LISA~\citep{lai2024lisa}/GLaMM~\citep{rasheed2024glamm}-style
\texttt{[SEG]} tokens: Kestrel~\citep{ahmed2025kestrel} wraps part
references in \texttt{<p>\dots</p>}/\texttt{[SEG]} and attaches
a segmentation decoder with query refinement to a
Uni3D-g~\citep{zhou2023uni3d} backbone, trained on the 3DCoMPaT-GrIn
split it constructs on
3DCoMPaT++~\citep{slim20253dcompat++}; PARIS3D~\citep{kareem2024paris3d} renders to
multi-view 2D for a ViT-H (SAM) backbone and routes
\texttt{[SEG]} through a custom mask decoder, releasing
RPSeg3D; SegPoint~\citep{he2024segpoint} adopts the same LISA-style
recipe for unified scene-level 3D segmentation and serves as
a part-task baseline in \citet{ahmed2025kestrel}. Box-based
Part-X-MLLM~\citep{wang2025part} pairs a dual-pathway encoder
(XYZ$+$normals; RGB) with a bounding-box grammar
($\langle\mathtt{boxs}\rangle\dots\langle\mathtt{boxe}\rangle$
plus \texttt{<adds>/<dels>/<mods>} edits) emitting parts as
autoregressive box tokens, with part ambiguity resolved by
post-hoc clustering of boxes through text semantics. All
inflate the non-LLM parameter budget by roughly an order of
magnitude, and in none does the LLM read or write a part as a
first-class vocabulary entry: Kestrel's \texttt{[SEG]} is a
single anchor slot re-used across all parts and decoded by an
external head, and Part-X-MLLM's box tokens address coordinate
bins rather than per-part identifiers. 3D-PLOT-LLM removes the
downstream machinery entirely and instead reserves
per-slot vocabulary tokens
\partk{0}\dots\partk{K{-}1}, addressable uniformly in prompts
and outputs and anchored to a fixed coarse spatial partition,
so the LLM learns slot-to-caption bindings end-to-end
without a segmentation or box head.

\paragraph{Scene-level 3D MLLMs and identifier-token precedent.}
A parallel line targets 3D MLLMs at the room-scene level
(3D-LLM~\citep{hong20233d}, LL3DA~\citep{chen2024ll3da},
LLaVA-3D~\citep{zhu2024llava}, 3D-LLaVA~\citep{deng20253d},
Reason3D~\citep{huang2025reason3d}, SpatialLM~\citep{mao2025spatiallm},
Pts3D-LLM~\citep{thomas2025pts3d}). A sub-thread
within it introduces learnable identifier tokens
as scene-level addressing:
Chat-Scene~\citep{huang2024chat} inserts per-object
\texttt{<OBJ\_k>} into the tokenizer; Descrip3D~\citep{xue2026descrip3d}
adds relational descriptions on top;
3DGraphLLM~\citep{zemskova20253dgraphllm} enriches each \texttt{<OBJ\_k>}
with a $k$NN subgraph; Grounded 3D-LLM~\citep{chen2024grounded} uses
a single \texttt{<ref>} delimited by \texttt{<p>...</p>}, trained
with a phrase-level contrastive CLASP objective. In 2D, Groma~\citep{ma2024groma} and Set-of-Mark prompting~\citep{yang2023set} bind identifiers to regions from a region proposer or an off-the-shelf segmenter. These works
establish vocabulary-level addressing as an effective
scene-level interface where an external proposer
(e.g.\ Mask3D~\citep{schult2023mask3d}) first carves the scene into
discrete objects; in every case the identifier is bound to a detected proposal. Applying this interface to parts within a
single object is qualitatively different because neither
enabling assumption holds: (i)~no external proposer produces
canonical part boundaries within an object (motivating our
deterministic, training-free partition, \S\ref{sec:bcp});
(ii)~while spatial disambiguation is a shared requirement at
both scene and part levels, parts within one object are often
semantically similar (e.g., four chair legs) and share the
same object-level bounding box, so the per-token global cues
that scene-level methods like Chat-Scene inject collapse to a
constant at the sub-object level (motivating our per-region
statistics within the object frame and inter-region adjacency
in \S\ref{sec:refinement}). MSR (\S\ref{sec:refinement}) is what lets the LLM ground language on regions that no detector has labeled.

\paragraph{Complementary part-level work.}
A separate thread targets part generation:
PartCrafter~\citep{lin2025partcrafter}, OmniPart~\citep{yang2025omnipart},
MMPart~\citep{bonakdar2025mmpart} produce decomposed part meshes via
image-conditioned generation. Outside the MLLM paradigm,
PartNet~\citep{mo2019partnet} provides a hierarchical 3D part
benchmark across 24 categories; PartSLIP~\citep{liu2023partslip} and
SATR~\citep{abdelreheem2023satr} transfer 2D vision-language
knowledge from GLIP~\citep{li2022grounded} to 3D part segmentation
via multi-view rendering; PartVerse~\citep{dong2025one} provides mesh-level part
annotations on Objaverse, which we adapt for Stage-2
PartVerse-QA supervision. These works produce geometry, masks, or
annotations rather than language-level part reasoning, and
complement ours.

\section{Method}
\label{sec:method}

\begin{figure}[t]
  \centering
  \includegraphics[width=0.95\linewidth]{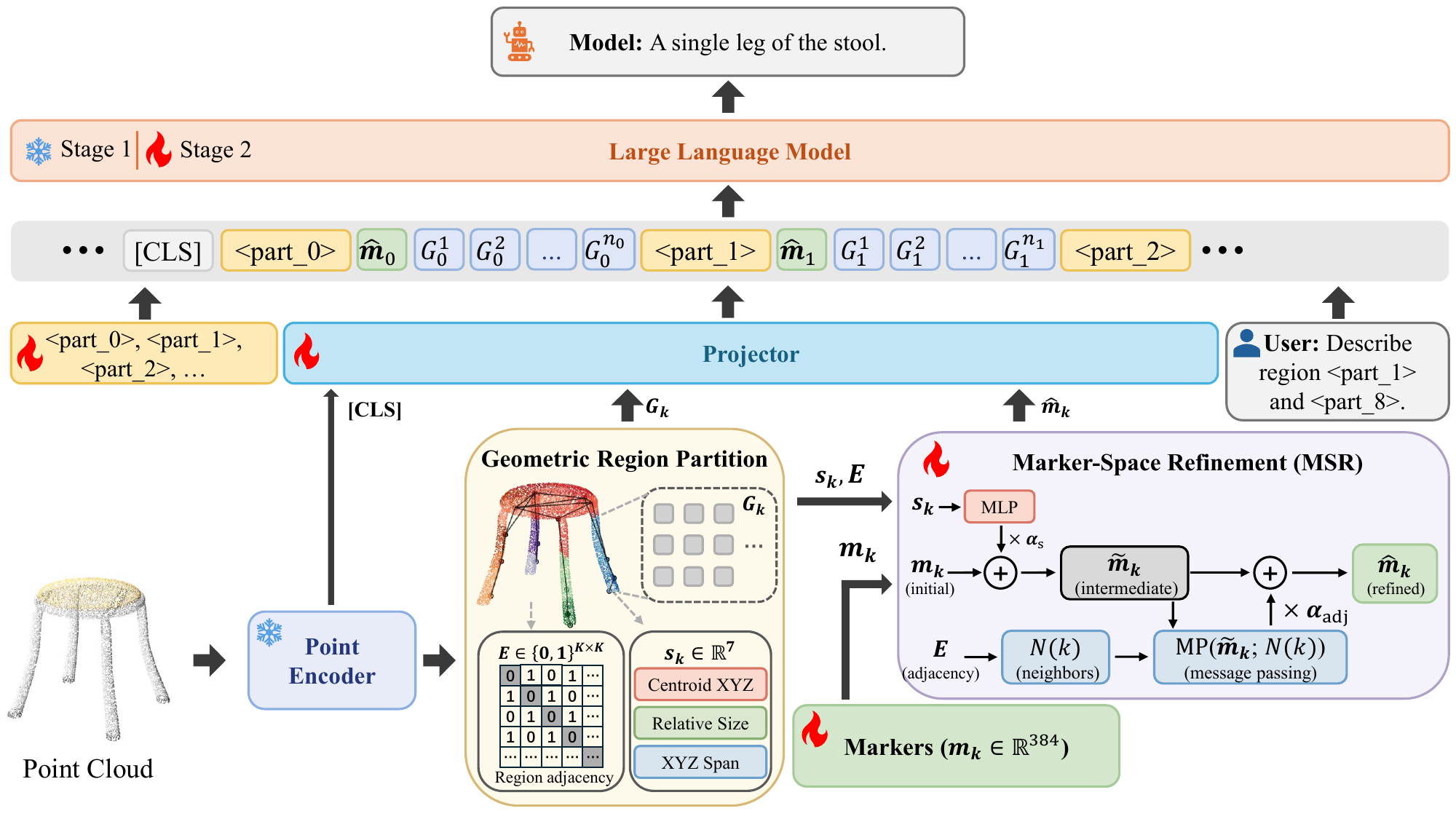}
  \caption{\textbf{3D-PLOT-LLM pipeline.} A frozen point
  encoder produces $512$ patch tokens that flow through three
  components: (i)~\textbf{Geometric Region Partition}
  (\S\ref{sec:bcp}) groups patches into $K$ regions $\{G_k\}$
  and exposes per-region statistics $s_k\!\in\!\R^7$ and
  inter-region adjacency $E$;
  (ii)~\textbf{Marker-Space Refinement}
  (\S\ref{sec:refinement}) updates each learnable marker
  $\mathbf{m}_k \to \hat{\mathbf{m}}_k$ via stats and graph
  residuals; (iii)~\textbf{Token Assembly}
  (\S\ref{sec:assembly}) interleaves \texttt{[CLS]}, reserved
  \partk{k} tokens, $\hat{\mathbf{m}}_k$, and $G_k$ before the
  LLM (frozen Stage~1, tuned Stage~2). \partk{k} is a
  vocabulary token and bypasses the projector; continuous
  tokens are projected. No segmentation decoder, bounding-box
  head, or point-wise mask supervision is used.}
  \label{fig:pipeline}
\end{figure}

\paragraph{Overview.}
3D-PLOT-LLM represents a 3D object as a structured token
sequence in which parts are first-class, addressable entries in
the LLM's vocabulary (Figure~\ref{fig:pipeline}). A frozen
point encoder tokenizes $8{,}192$ points into $512$ patch
tokens $\{x_i\}$ with centers $\{c_i\}$ and a global
\texttt{[CLS]}. Three components then operate before the LLM:
(i)~a deterministic, training-free region partition
(\S\ref{sec:bcp}) groups the patches into $K$ spatially
coherent regions and exposes per-region statistics and
inter-region adjacency;
(ii)~Marker-Space Refinement (MSR, \S\ref{sec:refinement})
conditions each region's learnable marker on those
statistics and adjacency; and
(iii)~per-region token assembly (\S\ref{sec:assembly})
inserts, before each region's patches, the refined marker
and a reserved vocabulary token~\partk{k} addressable in
both prompts and responses. No segmentation decoder,
bounding-box head, or point-wise mask supervision is used.

\subsection{Geometric region partition}
\label{sec:bcp}

We organize the $512$ patch tokens into $K$ spatially
coherent regions via a deterministic, training-free partition of the frozen encoder's patches (full algorithm in
App.~\ref{app:bcp}). The partition is not purely spatial: regions grow under a mixed cost of patch distance and feature dissimilarity, with weight $\lambda{=}0.8$ on distance (App.~\ref{app:bcp}). The partition is structural substrate,
not a semantic claim; it provides $K$ stable
addressable slots so that the per-region markers
(\S\ref{sec:assembly}), reserved vocabulary tokens \partk{k}, and MSR
(\S\ref{sec:refinement}) all index into a consistent structural
level. Two properties are required. \textbf{Determinism}: because
the encoder is frozen, the same input always yields the same
partition, so region index $k$ has a stable meaning across
inference runs and the LLM can learn slot-to-content bindings.
\textbf{Approximate size balance}: no single region should
carry a disproportionate share of the object's content, so
per-region markers' representational load remains roughly
comparable across $k$. Any spatially coherent partition with
these two properties would serve. For each region $k$ we record its
inter-region adjacency $E$ (induced by the patch
connectivity graph; App.~\ref{app:bcp}) and a 7-dim
statistics vector $s_k \in \R^7$ (region centroid $xyz$,
relative size, $xyz$ span); both feed into
MSR (\S\ref{sec:refinement}).

\subsection{Token assembly}
\label{sec:assembly}

For each region $k$, the assembly places a reserved label
token~\partk{k} and a learnable per-region marker
$\mathbf{m}_k \in \R^{384}$ immediately before that region's
patch tokens:
\[
\underbrace{\texttt{[CLS]}}_{\text{global}},\,
\underbrace{\partk{0}, \mathbf{m}_0, G_0}_{\text{region}~0},\,
\underbrace{\partk{1}, \mathbf{m}_1, G_1}_{\text{region}~1},\,\dots,\,
\underbrace{\partk{K{-}1}, \mathbf{m}_{K{-}1}, G_{K{-}1}}_{\text{region}~K{-}1}
\]
where $G_k$ is the set of patch tokens in region $k$,
serialized in their original encoder patch index order
inside the LLM input. The total LLM-input length is
$1{+}2K{+}\sum_k |G_k| = 545$ tokens at $K{=}16$. The tokens
in this assembly play two roles. The patch set $G_k$ and the
per-region marker $\mathbf{m}_k$ (a learnable $384$-d vector
in the projector's input space; the projector is a multi-layer
MLP mapping backbone features to the LLM's $4096$-d embedding)
form the \textbf{content} of each region. The reserved token \partk{k} (a 4096-d vocabulary
entry the LLM both reads and emits) acts as the \textbf{label};
it carries no intrinsic geometric content, with its meaning
arising entirely from the learned binding to its region's
content tokens. We therefore call \partk{k} a region pointer: it names a spatial region, not a category; the same index refers to different content on different objects; and unlike a closed-set semantic part vocabulary it composes with free-form language in both directions.

\subsection{Structural refinement}
\label{sec:refinement}

The per-region markers $\{\mathbf{m}_k\}$ introduced in
\S\ref{sec:assembly} are unaware of their region-level
spatial context: marker $\mathbf{m}_k$ does not see its
region's centroid or size, nor that it is adjacent to region
$k'$. Marker-Space Refinement (MSR) supplies this context
through a lightweight two-stage residual update on the
markers.

Let $s_k \in \R^7$ be region $k$'s statistics vector
($3$-dim centroid $xyz$ in the object frame, $1$-dim relative
size $|G_k| / N$ where $N{=}512$ is the patch budget, and
$3$-dim axis-aligned $xyz$ span), and
$\mathcal{N}(k) = \{k' : (k, k') \in E\}$ the set of regions
adjacent to $k$ (\S\ref{sec:bcp}). MSR updates each marker
$\mathbf{m}_k \in \R^{384}$ through two residuals:
\begin{align}
    \tilde{\mathbf{m}}_k &= \mathbf{m}_k + \alpha_{\text{s}} \cdot \text{MLP}(s_k), \label{eq:stats} \\
    \hat{\mathbf{m}}_k &= \tilde{\mathbf{m}}_k + \alpha_{\text{adj}} \cdot \text{MP}(\tilde{\mathbf{m}}_k; \mathcal{N}(k)), \label{eq:graph}
\end{align}
with $\alpha_{\text{s}}, \alpha_{\text{adj}}$ learnable residual
scales. $\text{MLP}: \R^7 \to \R^{384}$ has two linear
layers with hidden width $384$ and GELU activation in between.
$\text{MP}$ is a message-passing layer in the MPNN
sense~\citep{gilmer2017neural} with mean aggregation followed by a two-layer
update MLP $\phi$:
\begin{equation}
    \text{MP}(\tilde{\mathbf{m}}_k; \mathcal{N}(k)) =
        \phi\!\left(
            W_c \tilde{\mathbf{m}}_k +
            \frac{1}{|\mathcal{N}(k)|}
            \sum_{k' \in \mathcal{N}(k)} W_n \tilde{\mathbf{m}}_{k'}
        \right),
        \label{eq:mp}
\end{equation}
where $W_c, W_n \in \R^{384 \times 384}$ project the central
and neighbor markers respectively.
The residual form, combined with $\alpha_{\text{s}},
\alpha_{\text{adj}}$ initialized at $0.05$ and a zero-initialized
output layer in $\phi$, makes MSR start from identity
($\hat{\mathbf{m}}_k \approx \mathbf{m}_k$) at the beginning of
training. Branch contributions are quantified by a
stats-vs.-adjacency ablation in App.~\ref{app:msr-branch}.

The refined marker $\hat{\mathbf{m}}_k$ enters the projector
in place of $\mathbf{m}_k$. Because $\hat{\mathbf{m}}_k$ sits
immediately before $G_k$ in the assembly, the region-specific
structural context it carries directly conditions the LLM's
early attention over $G_k$; the patch tokens themselves remain
frozen encoder outputs.

\subsection{Training data and two-stage protocol}
\label{sec:training}

Training is alignment followed by instruction tuning,
matching the standard 3D MLLM
schedule~\citep{xu2024pointllm,qi2024shapellm}.
\textbf{Stage 1 (alignment)} trains the projector,
per-region markers, and MSR on 660K Cap3D captions of
Objaverse objects for 3~epochs, with the LLM and point
encoder frozen. \partk{k} tokens are deferred to Stage 2
because (i)~Cap3D captions are whole-object descriptions
with no part-level signal to learn their embeddings from,
and (ii)~the frozen Stage-1 LLM does not update its
input-embedding matrix in any case.
\textbf{Stage 2 (instruction tuning)} adds the \partk{k}
tokens to the tokenizer and trains the projector, per-region
markers, MSR, and the LLM on the union of
PointLLM's 70K complex instructions and 77K PartVerse-QA
pairs we derive from PartVerse mesh annotations
(App.~\ref{app:partverse}) for 3~epochs; the encoder remains
frozen. The PartVerse-QA
pairs cover two directions of the \partk{k} interface:
\textbf{caption-to-slots} (C2S; given a free-text part
caption, predict the \partk{k} set covering it) and
\textbf{slot-to-caption} (S2C; given a \partk{k} set,
generate a part caption). Together they supply the
part-grounded supervision under which \partk{k} embeddings
acquire bindings to region content. In matched-data
experiments without PartVerse-QA pairs (no-PV), the
\partk{k} embeddings receive no binding signal and the
C2S/S2C interface is non-functional; whole-object and
grounded-description tasks still benefit
(Tables~\ref{tab:main-obj},~\ref{tab:compat-papgd}).
Full hyperparameters in App.~\ref{app:hparams}.

\paragraph{Implementation.}
We instantiate the point encoder as
Point-BERT~\citep{yu2022point} pretrained via
ULIP-2~\citep{xue2024ulip}, and the LLM as
Vicuna-v1.5-7B~\citep{chiang2023vicuna}, matching
PointLLM~\citep{xu2024pointllm}'s released setup. Both
encoder and LLM are held identical across our variants and
the PointLLM baseline, so any gain is attributable to our
architectural contributions rather than encoder/LLM
strength. We use $K{=}16$ regions by default, set by the $512$-patch budget ($32$ patches per region) and motivated by
the PartVerse part-count distribution
(App.~\ref{app:partverse}); App.~\ref{app:sensitivity} studies the sensitivity to $K$ with a full $K{=}8$ retrain.

\section{Experiments}
\label{sec:exp}

\subsection{Evaluation protocols}
\label{sec:eval-protocols}

\paragraph{Language metrics.}
We report semantic metrics (SBERT~\citep{reimers2019sentence},
SimCSE~\citep{gao2021simcse}, GPT-4o LLM-as-judge) and
lexical-overlap metrics (BLEU-1~\citep{papineni2002bleu},
ROUGE-L~\citep{lin2004rouge}, METEOR~\citep{banerjee2005meteor}),
matching PointLLM~\citep{xu2024pointllm}; both families are
reported because lexical overlap alone underestimates
open-vocabulary captioning quality.
Three splits are evaluated: the $200$ Objaverse held-out
objects of the PointLLM captioning
benchmark~\citep{xu2024pointllm}; our held-out PartVerse-QA split, which
we construct on top of PartVerse~\citep{dong2025one} mesh
annotations (App.~\ref{app:partverse}); and 3DCoMPaT-GrIn
PaPGD's public test split~\citep{ahmed2025kestrel} ($6770$ multi-part
queries). The PartVerse-QA benchmark
(\S\ref{sec:training}) is scored as: \textbf{C2S} by Jaccard
and Exact-match over $392$ queries; \textbf{S2C} by
captioning metrics and a GPT-4o judge over $196$ queries. Objaverse traditional metrics
and PartVerse-QA S2C metrics (traditional and GPT-4o judge)
are all reported as $5$-run means under matched protocols;
per-run standard deviations for the main metrics are listed in
App.~\ref{app:variance}. C2S Jaccard and Exact-match use
deterministic decoding (single run); 3DCoMPaT-GrIn
(Table~\ref{tab:compat-papgd}) follows cited-baseline
reporting.

\subsection{Main results}
\label{sec:main-results}

\paragraph{Objaverse captioning.}
We report two model variants in Table~\ref{tab:main-obj} to
disentangle architectural and data contributions: a
matched-data variant (no-PV, trained on the same
$660$K Cap3D + $70$K complex data as PointLLM) and our full
model (adding $77$K of our PartVerse-QA pairs at Stage~2).
All six metrics (SBERT, SimCSE, GPT-4o judge, BLEU-1,
ROUGE-L, METEOR) are reported as $5$-run means; per-cell
std in Table~\ref{tab:obj-variance} (App.~\ref{app:variance}).
Under matched data, 3D-PLOT-LLM no-PV outperforms PointLLM
on all six metrics (up to $+1.48$ GPT-4o judge), an
architectural-level lift on whole-object captioning.
With PartVerse-QA at Stage~2, the gain extends to $+0.65$
SBERT and $+1.85$ GPT-4o judge over PointLLM, leading PointLLM on
all six metrics; METEOR ($+0.40$) reaches significance
($p\!\approx\!0.04$). The other two re-evaluated baselines, ShapeLLM and ENEL~\citep{tang2026exploring}, score higher than PointLLM on the lexical-overlap metrics but not on the semantic ones: ShapeLLM is far below on SBERT, SimCSE, and the GPT-4o judge, and ENEL is within noise of PointLLM on SBERT and SimCSE and below it on the judge ($37.02$ vs $39.08$).

We further benchmark against PointLLM-PiSA~\citep{guo2026pisa},
which adds $62$K self-augmented captions ($132$K total)
targeting captioning quality. Our $147$K-total mix targets
part grounding instead and leads PiSA on $4$ of $5$
traditional metrics (SBERT, SimCSE, BLEU-1, METEOR; PiSA's
GPT-4o cell is ``--'' due to its closed-source judge
pipeline). Part-grounded supervision is therefore a viable
scaling axis complementary to whole-object caption
augmentation.

\begin{table}[t]
  \caption{\textbf{Objaverse captioning.} 3D-PLOT-LLM (full)
  leads SBERT, SimCSE, GPT-4o; the matched-data no-PV variant
  outperforms PointLLM on all six metrics at the same $70$K
  Stage-2 budget, isolating architectural from data-volume
  contribution. Cells are $5$-run means; per-cell $\pm$std
  and significance in App.~\ref{app:variance}. Stage-2 reports total instruction-tuning samples.
  \textbf{Best} and \underline{2nd} per column.}
  \label{tab:main-obj}
  \centering\small
  \renewcommand{\arraystretch}{0.9}
  \setlength{\tabcolsep}{3pt}
  \begin{tabular}{llcccccc}
    \toprule
    Method & Stage-2 & SBERT$\uparrow$ & SimCSE$\uparrow$ & GPT-4o$\uparrow$ & BLEU-1$\uparrow$ & ROUGE-L$\uparrow$ & METEOR$\uparrow$ \\
    \midrule
    PointLLM-PiSA              & $132$K & $\underline{48.21}$ & $48.38$              & --                  & $3.81$              & $7.29$  & $12.32$             \\
    ShapeLLM (re-eval)         & $75$K  & $35.49$             & $36.13$              & $12.86$             & $\underline{6.02}$     & $\underline{8.00}$     & $12.33$ \\
    ENEL (re-eval)             & $70$K  & $47.70$             & $48.22$              & $37.02$             & $\mathbf{15.18}$    & $\mathbf{19.28}$    & $\mathbf{17.89}$ \\
    PointLLM (re-eval)         & $70$K  & $47.66$             & $48.16$              & $39.08$             & $3.80$              & $7.02$              & $12.19$             \\
    3D-PLOT-LLM (no-PV)        & $70$K  & $47.94$             & $\underline{48.44}$  & $\underline{40.56}$ & $3.84$              & $7.10$              & $12.27$             \\
    \textbf{3D-PLOT-LLM (full)} & $147$K & $\mathbf{48.31}$   & $\mathbf{48.67}$     & $\mathbf{40.93}$    & $3.94$  & $7.22$              & $\underline{12.59}$    \\
    \bottomrule
  \end{tabular}
\end{table}

\paragraph{PartVerse-QA.}
PartVerse-QA probes a capability no prior part-aware
3D MLLM natively supports (\S\ref{sec:related}): bidirectional
\partk{k}-token addressing in a single LLM token stream.
Prior methods emit part references as decoder masks or boxes
but accept no region encoding as model input, so direct
cross-method comparison is structurally constrained
(App.~\ref{app:partverse}, ``Cross-method evaluation''); the
architectural ranking is independently validated against
partition-agnostic mesh GT in App.~\ref{app:mesh-miou}.
Table~\ref{tab:main-partverse} compares our own architectural
progression under identical PartVerse-QA supervision. 3D-PLOT-LLM reaches
Jaccard $0.459$ and Exact-match $13.78\%$ on the held-out
split (${\sim}77\!\times$ random chance at the median query
cardinality of $3$ over $K{=}16$ slots; App.~\ref{app:partverse}). Vocabulary addressing alone (vocab-only) already
establishes a non-trivial $0.413$ Jaccard / $8.42\%$
Exact-match baseline; MSR then lifts both metrics
($+0.046$ Jaccard; $+5.36$ percentage points / $+64\%$
relative on Exact-match). The intermediate variant with raw
markers but no MSR (row~2) regresses below vocab-only
($0.263$ Jaccard, $3.06\%$ Exact-match): un-conditioned
markers carry no region-level structural signal
(\S\ref{sec:refinement}), distracting early attention
without informing it; only MSR's conditioning on per-region
statistics and adjacency makes the marker substrate functional.
Our model also leads every S2C metric (SBERT $62.04$, GPT-4o
$44.68$, METEOR $34.21$), with the GPT-4o judge gap
significant ($+2.75$, $p\!\approx\!0.012$;
App.~\ref{app:variance}). A label-permutation probe
(App.~\ref{app:permutation}) collapses Jaccard from
$0.459$ to $0.209$ and Exact-match from $13.78\%$ to
$1.28\%$ when \partk{k} identifiers are shuffled at
inference but markers and patches are unchanged, evidence
that the LLM has learned a genuine token-region binding
rather than treating \partk{k} as a cosmetic label.
App.~\ref{app:find3d} projects an open-vocabulary part segmenter, Find3D~\citep{ma2025find}, onto the slot grid under the identical protocol (the part-aware 3D MLLMs of \S\ref{sec:related} either have no public checkpoints or take rendered views or scene-level inputs): it reaches Jaccard $0.304$ against our $0.459$ and trails in every mapping-quality stratum.

\begin{table}[t]
  \caption{\textbf{Architectural progression on PartVerse-QA.}
  Vocab-only is the no-marker reference; raw markers without
  MSR (row~2) lack region-level context and regress below
  vocab-only (\S\ref{sec:refinement}). MSR reaches Jaccard
  $0.459$ and Exact-match $13.78\%$ ($+64\%$ relative over
  vocab-only) at $0.7$M extra parameters. C2S uses
  deterministic decoding; S2C metrics are $5$-run means
  (App.~\ref{app:variance}). Alternative refinement designs
  in App.~\ref{app:refinement-alt}.}
  \label{tab:main-partverse}
  \centering\small
  \renewcommand{\arraystretch}{0.85}
  \setlength{\tabcolsep}{4pt}
  \begin{tabular}{lcccccc}
    \toprule
    & \multicolumn{2}{c}{Caption-to-slots} & \multicolumn{4}{c}{Slot-to-caption} \\
    \cmidrule(lr){2-3} \cmidrule(lr){4-7}
    Method & Jaccard$\uparrow$ & Exact-match$\uparrow$ & SBERT$\uparrow$ & GPT-4o$\uparrow$ & METEOR$\uparrow$ & Word-F1$\uparrow$ \\
    \midrule
    Vocab-only            & $\underline{0.413}$  & $\underline{8.42\%}$  & $61.23$              & $\underline{43.98}$ & $33.72$              & $0.416$             \\
    $+$ Markers (no MSR)  & $0.263$              & $3.06\%$              & $\underline{61.57}$  & $41.93$             & $\underline{33.99}$  & $\underline{0.420}$ \\
    \textbf{$+$ MSR (ours)} & $\mathbf{0.459}$   & $\mathbf{13.78\%}$    & $\mathbf{62.04}$     & $\mathbf{44.68}$    & $\mathbf{34.21}$     & $\mathbf{0.421}$    \\
    \bottomrule
  \end{tabular}
\end{table}

\paragraph{3DCoMPaT-GrIn part-grounded description, text track.}
3DCoMPaT-GrIn PaPGD (Table~\ref{tab:compat-papgd}) is the
part-aware point grounded description split of 3DCoMPaT++
\citep{slim20253dcompat++,ahmed2025kestrel}. We fine-tune both
PointLLM (``our FT'' row) and 3D-PLOT-LLM on the
3DCoMPaT-GrIn training split for this evaluation; our Stage-2
here excludes PartVerse-QA, adding no benchmark-related
supervision over the cited baselines, so the gap reflects the
architectural contribution. The benchmark is
originally a joint segmentation$+$text task; as 3D-PLOT-LLM
is decoder-free, we evaluate the text-description track here
(App.~\ref{app:prompts} (f)) and the grounding track separately below. Only the PointLLM and 3D-PLOT-LLM rows are run by us (hence their GPT-4o cells); the other rows are cited. 3D-PLOT-LLM outperforms
PointLLM on all five reported text metrics ($+0.47$ BLEU-4,
$+0.74$ METEOR, $+0.81$ SBERT, $+0.90$ SimCSE, $+3.03$
GPT-4o judge). It also leads PARIS3D,
SegPoint, and Kestrel on every reported metric, and ShapeLLM
on three of four ($+1.31$ BLEU-4, $+3.88$ METEOR, $+0.85$
SimCSE). The
semantic GPT-4o judge widens the gap relative to lexical
and embedding metrics, mirroring the amplification on
PartVerse-QA S2C.

\begin{table}[t]
  \caption{\textbf{3DCoMPaT-GrIn PaPGD, multi-part split, text-description metrics.}
  3D-PLOT-LLM outperforms PointLLM, Kestrel, PARIS3D, and
  SegPoint on every reported text metric, and ShapeLLM on $3$ of $4$.
  Cited baseline rows are from Kestrel's Table~1; our row
  excludes PartVerse-QA from Stage-2 to ensure fair comparison.}
  \label{tab:compat-papgd}
  \centering\small
  \renewcommand{\arraystretch}{0.85}
  \setlength{\tabcolsep}{5pt}
  \begin{tabular}{lccccc}
    \toprule
    Method & BLEU-4$\uparrow$ & METEOR$\uparrow$ & SBERT$\uparrow$ & SimCSE$\uparrow$ & GPT-4o$\uparrow$ \\
    \midrule
    ShapeLLM                                & $8.01$              & $30.05$              & $\mathbf{84.74}$    & $\underline{85.38}$  & --                  \\
    PARIS3D                                 & $8.45$              & $31.62$              & $82.08$             & $83.86$              & --                  \\
    SegPoint                                & $7.24$              & $27.62$              & $78.50$             & $81.47$              & --                  \\
    Kestrel                                 & $8.55$              & $32.88$              & $82.19$             & $84.13$              & --                  \\
    PointLLM (our FT)                       & $\underline{8.85}$  & $\underline{33.19}$  & $83.43$             & $85.33$              & $\underline{40.31}$ \\
    \textbf{3D-PLOT-LLM (ours)}        & $\mathbf{9.32}$     & $\mathbf{33.93}$     & $\underline{84.24}$ & $\mathbf{86.23}$     & $\mathbf{43.34}$    \\
    \bottomrule
  \end{tabular}
\end{table}

\paragraph{3DCoMPaT-GrIn part-grounded description, grounding track.}
Each slot is the union of its patches' points, so the interface also induces point masks. We issue every GrIn part name as a caption-to-slots query to the model trained on PartVerse-QA, which receives no GrIn training, and score the induced mask by mIoU against per-point fine part labels. It reaches $0.221$ / $0.220$ / $0.242$ on the single-part / multiple-parts / embodied splits, against oracle ceilings of $0.448$ / $0.441$ / $0.488$ for the best slot set on the $K{=}16$ grid (App.~\ref{app:grin-grounded}). This is about half of what the interface can express without a decoder; slots are addressable coarse regions, not boundary-accurate masks.

\paragraph{Parameter efficiency.}
3D-PLOT-LLM adds $\approx\!0.8$M architectural parameters
(markers $6$K, \partk{k} embeddings $66$K, MSR
$\approx\!0.74$M) on top of the shared frozen $22$M encoder
and fine-tuned LLM, an order of magnitude below the
segmentation-decoder and box-grammar machinery of prior
part-aware 3D MLLMs.

\subsection{Ablations}
\label{sec:ablation}

\paragraph{Component ablation.}
Table~\ref{tab:component-ablation} isolates the four
architectural pieces: $K$-slot grouping, per-region markers,
reserved \partk{k} tokens (training-coupled with PartVerse-QA),
and MSR. Input token reorganization alone yields $+0.40$
SBERT and $+1.60$ GPT-4o over the unstructured-sequence
PointLLM baseline; adding markers extends the
SBERT gain to $+0.55$. Vocab
tokens alone (row 4) establish a non-trivial $0.413$ C2S
Jaccard and $43.98$ S2C GPT-4o grounding signal; MSR's
conditioning on per-region stats and adjacency makes the
marker substrate functional, lifting C2S Jaccard to $0.459$,
S2C GPT-4o to $44.68$, and S2C SBERT to $62.04$ (row 6, ours)
at only $0.7$M extra parameters. Inserting the interface does not degrade whole-object captioning: every interface variant is at or above PointLLM on the Objaverse GPT-4o judge ($40.56$ to $41.36$ vs $39.08$; App.~\ref{app:variance}).

\begin{table}[t]
  \caption{\textbf{Component ablation.} MSR (row~6, ours)
  leads on every PartVerse-QA metric (C2S, S2C-SB, S2C-GPT)
  and on Objaverse SBERT. Grp: grouping; Mkr: markers;
  L+PV: reserved \partk{k}$+$PartVerse-QA; C2S:
  caption-to-slots Jaccard; S2C-SB / S2C-GPT:
  slot-to-caption SBERT / GPT-4o judge.}
  \label{tab:component-ablation}
  \centering\small
  \renewcommand{\arraystretch}{0.85}
  \setlength{\tabcolsep}{3pt}
  \begin{tabular}{lccccccccc}
    \toprule
    & \multicolumn{4}{c}{Components} & \multicolumn{2}{c}{Objaverse} & \multicolumn{3}{c}{PartVerse-QA} \\
    \cmidrule(lr){2-5} \cmidrule(lr){6-7} \cmidrule(lr){8-10}
    Method & Grp & Mkr & L+PV & MSR & SBERT$\uparrow$ & GPT-4o$\uparrow$ & C2S$\uparrow$ & S2C-SB$\uparrow$ & S2C-GPT$\uparrow$ \\
    \midrule
    PointLLM                & \ding{55} & \ding{55} & \ding{55} & \ding{55} & $47.66$            & $39.08$           & --                  & --                  & --                  \\
    $+$ Grouping            & \ding{51} & \ding{55} & \ding{55} & \ding{55} & $48.06$            & $40.68$           & --                  & --                  & --                  \\
    $+$ Markers             & \ding{51} & \ding{51} & \ding{55} & \ding{55} & $\underline{48.21}$ & $40.56$             & --                  & --                  & --                  \\
    $+$ L+PV (no Mkr)       & \ding{51} & \ding{55} & \ding{51} & \ding{55} & $47.83$             & $40.77$             & $\underline{0.413}$ & $61.23$             & $\underline{43.98}$ \\
    $+$ Mkr+L+PV (no MSR)   & \ding{51} & \ding{51} & \ding{51} & \ding{55} & $48.08$             & $\mathbf{41.36}$    & $0.263$             & $\underline{61.57}$ & $41.93$             \\
    \textbf{$+$ MSR (full)} & \ding{51} & \ding{51} & \ding{51} & \ding{51} & $\mathbf{48.31}$    & $\underline{40.93}$ & $\mathbf{0.459}$    & $\mathbf{62.04}$    & $\mathbf{44.68}$    \\
    \bottomrule
  \end{tabular}
\end{table}

\paragraph{Additional ablations.} Five supplementary ablations
support each design choice: PartVerse-QA data scaling shows
near-monotonic, steeply rising C2S and S2C curves
(App.~\ref{app:data-scaling}); refinement-location ablation
favors the marker substrate over \partk{k} embeddings
(App.~\ref{app:refinement-alt}); the MSR branch split
shows stats and adjacency carry complementary signal
($0.284$ and $0.305$ Jaccard alone, $0.459$ combined;
App.~\ref{app:msr-branch}); and App.~\ref{app:sensitivity} reports a full $K{=}8$ retrain (part description unchanged, S2C SBERT $62.30$ vs $62.04$; grounding point-IoU $0.460 \to 0.406$) and an inference-time partition sweep (at most $0.8$ SBERT and $0.013$ Jaccard of change).

\subsection{Qualitative results}
\label{sec:qualitative}

\begin{figure}[t]
  \centering
  \includegraphics[width=0.92\linewidth]{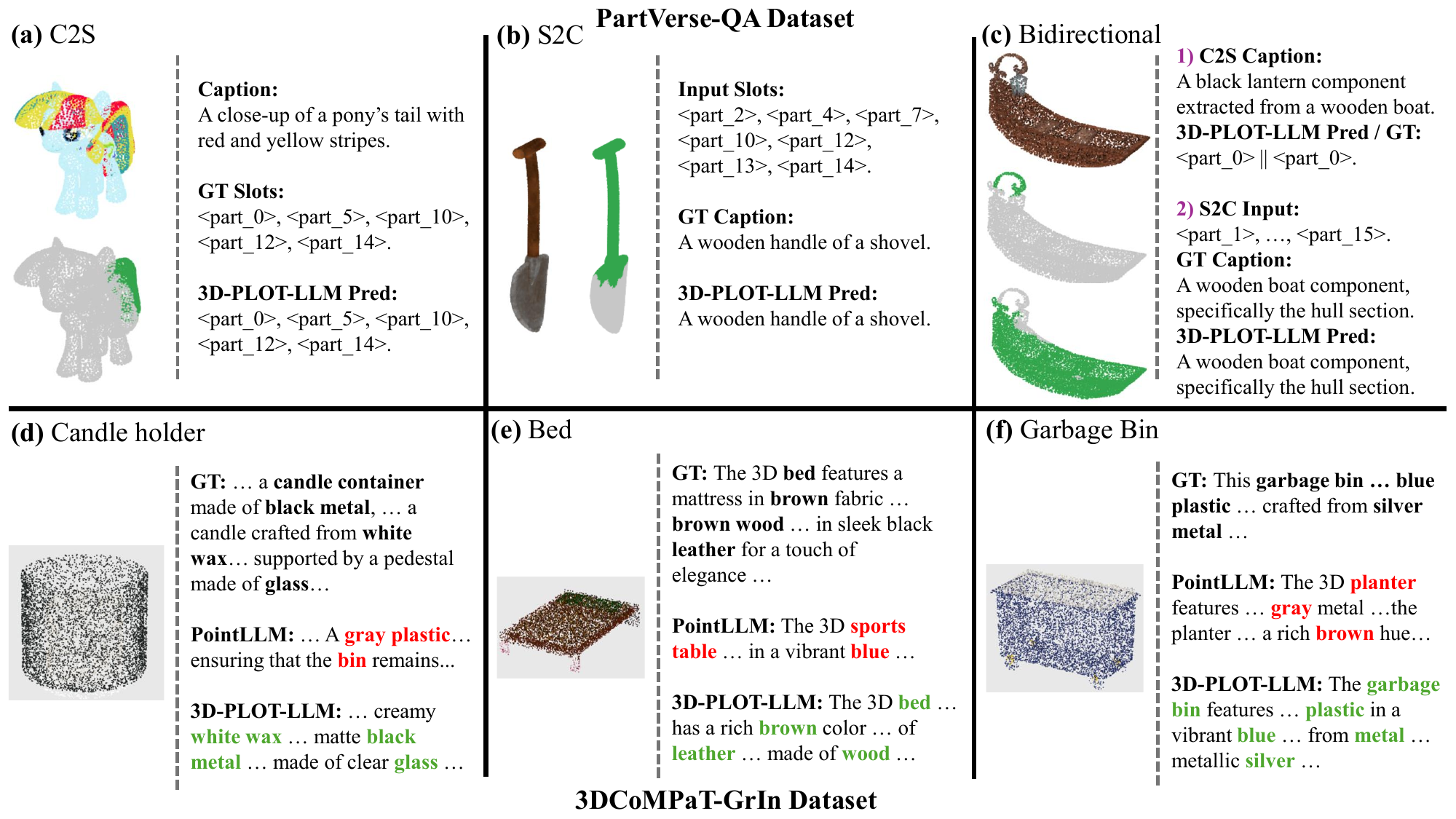}
  \caption{\textbf{Qualitative examples of part-aware
  tokenization.} \textbf{Top row (PartVerse-QA,
  vocabulary-level part addressing).} 3D-PLOT-LLM
  (a) maps a free-form caption to the exact \partk{k} set
  covering the described region (C2S, pony tail, $5$ slots,
  Jaccard $1.00$); (b) describes a multi-slot input in one
  sentence (S2C, shovel handle, $7$ slots
  $\to$ ``a wooden handle of a shovel'', Word-F1 $1.00$);
  (c) handles both directions on the same object across
  granularities ranging from a single-slot lantern to a
  near-whole-object $10$-slot hull. \textbf{Bottom row
  (3DCoMPaT-GrIn, part-aware grounded description).}
  PointLLM~\citep{xu2024pointllm} drifts to a wrong object
  class and confabulates parts to fit the misclassification
  (``bin'', ``sports table'', ``planter''; \textcolor{red}{red});
  3D-PLOT-LLM stays anchored to the correct class
  (candle holder, bed, garbage bin) and recovers fine-grained
  material/color attributes (\textcolor{green}{green}:
  matches GT). Full dialogues, additional examples, and
  failure modes in App.~\ref{app:qualitative}.}
  \label{fig:qualitative}
\end{figure}

Figure~\ref{fig:qualitative} illustrates two regimes.
\textbf{Top}: a single \partk{k} vocabulary carries both
C2S and S2C across granularities, with slot sets emerging
as data-learned compositions that correspond to no
predefined taxonomy. \textbf{Bottom}: PointLLM drifts to
wrong classes (candle holder $\to$ bin, bed $\to$ sports
table, garbage bin $\to$ planter) and confabulates parts to
fit the misclassification, while 3D-PLOT-LLM stays on-class
and recovers multi-part materials/colors, the qualitative
signature of the $+3.03$ GPT-4o gap in
Table~\ref{tab:compat-papgd}.

\section{Limitations and future work}
\label{sec:limits}

3D-PLOT-LLM addresses parts at $K$-slot granularity rather
than point-wise: ``where is the handle'' returns a slot set
of \partk{k} tokens, each a patch cluster, which suffices
for language-level part QA but is not a substitute for
segmentation-decoder methods such as
Kestrel~\citep{ahmed2025kestrel} when point-wise masks are
needed. Repeated parts are spatially distinct by construction (each wheel has its own slot; a collective reference grounds to their union), but a singular reference to one of several identical parts cannot pick an instance (Fig.~\ref{fig:appx-fail}c), while S2C captions a supplied wheel slot cleanly (Fig.~\ref{fig:appx-partverse-s2c}a): the ambiguity is on the query side. PartVerse-QA is built around the fixed $K{=}16$ interface, and its union-IoU filter keeps only the representable pairs ($73\%$ of part-annotated points; on the unfiltered strata the model still stays $2.0$ to $2.6\times$ above chance, App.~\ref{app:partverse}). Direct comparisons are limited to baselines with public checkpoints. The fixed slot budget $K{=}16$ accommodates
$94.5\%$ of objects in the aligned PartVerse pool by part
count (App.~\ref{app:partverse}); a
learned variable-$K$ or hierarchical partition is left to
future work, alongside swapping in a stronger backbone
(ReCon++~\citep{qi2024shapellm},
Uni3D~\citep{zhou2023uni3d}) without architectural change. For full scenes, our part tokens can be nested under object identifiers from scene-level machinery, since the partition needs only an object's patch tokens. C2S and S2C also compose without curated slot sets: in interactive use, the slot set given to an S2C query comes from the model's own C2S answer or from a point the user clicks, which maps deterministically to its slot. Composed part QA, such as asking for the material of the handle, and point-referred questions need instruction data that PartVerse's caption-style annotations do not supply; building such QA pairs is the next step. Code, checkpoints, the benchmark, and the partition packs are publicly available (link in \S\ref{sec:intro}).

\section{Conclusion}
\label{sec:conclusion}
We showed that an object-level 3D MLLM can be made
part-aware without a segmentation decoder or bounding-box
grammar, by turning every geometric region into a
first-class addressable token refined by Marker-Space
Refinement (MSR). 3D-PLOT-LLM introduces vocabulary-level
part addressing on PartVerse-QA, leads PointLLM, Kestrel, PARIS3D, and SegPoint on every text-description metric of 3DCoMPaT-GrIn
PaPGD, and lifts Objaverse captioning above PointLLM, all
with under $1$M additional parameters on a frozen point
encoder. A label-permutation probe (App.~\ref{app:permutation}) confirms
that the LLM learns a genuine token-region binding rather than
treating \partk{k} as a cosmetic label. Reframing addressable parts
as a vocabulary problem opens a lightweight, decoder-free
route to part-aware 3D language models, with part-grounded
supervision a viable scaling axis complementary to
whole-object caption augmentation.

\begin{ack}
The authors acknowledge the Center for Advanced Research Computing (CARC) at the University of Southern California for providing computing resources that have contributed to the research results reported within this publication (\url{https://carc.usc.edu}). This work received no external funding. The authors declare no competing interests.
\end{ack}

\bibliographystyle{plainnat}
\bibliography{references}

\appendix

\section{Geometric region partition: algorithm}
\label{app:bcp}

This appendix expands \S\ref{sec:bcp}. The partition procedure
takes the $N{=}512$ patch centers $\{c_i\}$ and patch features
$\{f_i\}$ from a frozen Point-BERT and returns a grouping into
$K{=}16$ regions. This specific procedure is implementation,
not contribution: any spatially coherent deterministic
partition with approximate size balance would serve the
architectural argument equally well.

The algorithm has four stages.

\textbf{(S0) Patch connectivity
graph.} A $k$-nearest-neighbor graph ($k{=}12$) over the
$512$ patch centers, with edges exceeding the 75th percentile
of $k$NN lengths gated to prevent thin regions from
short-circuiting across empty space. The gating can leave
the graph disconnected; for isolated regions (empty
$\mathcal{N}(k)$), the neighbor sum in Eq.~\ref{eq:mp} is
dropped and MP reduces to $\phi(W_c \tilde{\mathbf{m}}_k)$.

\textbf{(S1) Superpixel growth.}
Farthest-point seeds expand by adding neighbors in order of
a mixed cost
$c_{uv} = \lambda\,\|c_u - c_v\|_2 + (1-\lambda)(1 -
\cos(f_u, f_v))$ with $\lambda{=}0.8$. Each superpixel caps at
$\sim$3 patches, yielding $150$--$200$ superpixels.

\textbf{(S2) Balanced merge to $K$.}
On the superpixel adjacency graph, we pick $K{=}16$ geodesic
farthest-point seeds and grow each by priority-queue
expansion under the same cost $c_{uv}$ (lowest-cost adjacent
unassigned superpixel first), capped at
$\lceil 512 / K \rceil = 32$ patches per region. Disconnected
components seed and expand independently, with seeds
allocated in proportion to component size. The result is
regions of $\sim\!32$ patches on average.

\textbf{(S3) Boundary smoothing and connectivity cleanup.}
Four iterations reassign each boundary patch to the region
with lower mean $c_{uv}$ to its in-region neighbors (ties
broken by region index). A final connectivity pass splits
non-contiguous regions and absorbs fragments smaller than
four patches into the lowest-cost neighbor; any excess
regions are merged by centroid distance until exactly
$K{=}16$ regions remain. All $512$ patches end up assigned;
final region sizes may deviate from the S2 cap of $32$ due
to boundary reassignment and fragment absorption.

The partition is fully deterministic given the frozen
Point-BERT output (the random seed is derived from the patch
centers), so identical centers produce identical partitions;
we extract it once per object and cache it offline. The choice
of $K{=}16$ is motivated in App.~\ref{app:partverse}.

\section{Hyperparameters and compute}
\label{app:hparams}

All runs use $2\!\times\!\text{A100 80GB}$ GPUs with bf16 mixed
precision.
\textbf{Stage 1 (alignment):} LLM and Point-BERT frozen;
projector, per-region markers, and MSR trainable. Data: 660K Cap3D
captions of Objaverse, $3$ epochs, LR $2\!\times\!10^{-3}$ with
cosine schedule and 3\% warm-up; effective batch 48;
$\sim$48 hours per run.
\textbf{Stage 2 (instruction tuning):} full LLM, projector,
markers, and MSR all trainable. \partk{k} tokens are added to
the tokenizer at this stage, initialized to the mean of the
existing vocabulary embeddings, and trained jointly. Data:
PointLLM's 70K complex instructions $+$ 77K PartVerse-QA (\S\ref{sec:training}), $3$ epochs, LR
$2\!\times\!10^{-5}$ with cosine schedule and 3\% warm-up;
effective batch 16 under DeepSpeed
ZeRO-3~\citep{rasley2020deepspeed}; $\sim$40 hours per run.
\textbf{MSR:} both MLPs in MSR (the statistics MLP
$\text{MLP}_s : \R^7 \to \R^{384}$ in Eq.~\ref{eq:stats} and
the message MLP $\phi$ inside $\text{MP}(\cdot;\mathcal{N}(k))$
in Eqs.~\ref{eq:graph}--\ref{eq:mp}) use GELU activations
and no bias on either of their two linear layers. Residual scales
$\alpha_{\text{s}}, \alpha_{\text{adj}}$ of
Eqs.~\ref{eq:stats}--\ref{eq:graph} are initialized to $0.05$.
The output layer of $\phi$ in Eq.~\ref{eq:mp} (weight and bias)
is zero-initialized so that
$\hat{\mathbf{m}}_k \approx \mathbf{m}_k$ at the start of
training; $\text{MLP}_s$ uses Xavier initialization on its
two linear layers.
\textbf{Evaluation:} sampling-based inference scores are
averaged over $5$ seeds;
see \S\ref{sec:eval-protocols} for per-split details.

\section{Per-run variance and significance tests}
\label{app:variance}
This appendix backs the main-paper Tables~\ref{tab:main-obj}
(Objaverse) and~\ref{tab:main-partverse} (PartVerse-QA) with
$5$-run mean$\,\pm\,$std and pairwise Welch's two-sample
$t$-tests ($n_1{=}n_2{=}5$, two-sided). Each seed
independently re-runs sampling-based inference on the same
trained ckpt. PointLLM, ShapeLLM, and ENEL are re-evaluated by us
under the matched protocol, so their $5$-run std are reported
in Table~\ref{tab:obj-variance} alongside our variants;
PiSA (Table~\ref{tab:main-obj}) is cited from its paper and the ShapeLLM, PARIS3D, SegPoint, and Kestrel rows of Table~\ref{tab:compat-papgd} from Kestrel's Table~1; these sources give only published means or single-run
numbers without per-run variance estimates.

Excluded from std reporting: the cited PiSA row (no per-run
variance released); PartVerse-QA C2S Jaccard and C2S
Exact-match (deterministic decoding by design, reported in
main Table~\ref{tab:main-partverse} without std);
3DCoMPaT-GrIn (no per-run variance reported by cited
baselines). Table~\ref{tab:obj-variance} unifies the
variance panels: the four semantic / lightweight-lexical
Objaverse metrics (SBERT, SimCSE, GPT-4o, METEOR) and the
two PartVerse-QA S2C metrics with non-trivial $5$-run
variance (SimCSE and the GPT-4o judge); the
latter two columns are blank for rows whose model is not
trained against the \partk{k} part-token interface
(ShapeLLM, ENEL, PointLLM, no-PV).

\begin{table}[!ht]
  \caption{\textbf{Per-row mean$\,\pm\,$std (Objaverse +
  PartVerse-QA S2C).} All cells are $5$-run
  mean$\,\pm\,$std (Objaverse on val\_$200$, PartVerse-QA S2C
  on the $196$-query held-out split). ``---'': PartVerse-QA
  S2C metric not applicable to that row (ShapeLLM, ENEL, and PointLLM
  have no \partk{k} interface at all; no-PV has the interface
  architecturally but is not trained on PartVerse-QA, so its
  \partk{k} embeddings remain unbound). PartVerse-QA C2S Jaccard and
  Exact-match are deterministic (no std applicable) and are
  reported in main Table~\ref{tab:main-partverse}.
  Pairwise Welch's $t$-test results are summarized in the
  ``Pairwise significance tests'' paragraph below.
  \textbf{Best} and
  \underline{2nd} per column among rows with applicable
  data.}
  \label{tab:obj-variance}
  \centering\scriptsize
  \renewcommand{\arraystretch}{1.0}
  \setlength{\tabcolsep}{2pt}
  \begin{tabular}{lcccccc}
    \toprule
    & \multicolumn{4}{c}{Objaverse} & \multicolumn{2}{c}{PartVerse-QA S2C} \\
    \cmidrule(lr){2-5}\cmidrule(lr){6-7}
    Method & SBERT & SimCSE & GPT-4o & METEOR & SimCSE & GPT-4o \\
    \midrule
    ShapeLLM (re-eval)             & $35.49 {\pm} 0.37$              & $36.13 {\pm} 0.49$              & $12.86 {\pm} 1.35$              & $12.33 {\pm} 0.09$              & ---                                 & ---                                 \\
    ENEL (re-eval)                 & $47.70 {\pm} 0.69$              & $48.22 {\pm} 0.70$              & $37.02 {\pm} 1.29$              & $\mathbf{17.89 {\pm} 0.51}$     & ---                                 & ---                                 \\
    PointLLM (re-eval)             & $47.66 {\pm} 0.58$              & $48.16 {\pm} 0.75$              & $39.08 {\pm} 1.83$              & $12.19 {\pm} 0.30$              & ---                                 & ---                                 \\
    3D-PLOT-LLM (no-PV)            & $47.94 {\pm} 0.52$              & $48.44 {\pm} 0.50$              & $40.56 {\pm} 1.55$              & $12.27 {\pm} 0.14$              & ---                                 & ---                                 \\
    Vocab-only                     & $47.83 {\pm} 0.33$              & $48.56 {\pm} 0.29$              & $40.77 {\pm} 1.06$              & $12.42 {\pm} 0.52$              & $\underline{62.24 {\pm} 0.80}$      & $\underline{43.98 {\pm} 1.25}$      \\
    $+$ Markers (no MSR)           & $\underline{48.08 {\pm} 0.49}$  & $\mathbf{48.92 {\pm} 0.54}$     & $\mathbf{41.36 {\pm} 1.26}$     & $12.50 {\pm} 0.17$  & $62.23 {\pm} 0.92$                  & $41.93 {\pm} 1.56$                  \\
    \textbf{3D-PLOT-LLM (ours)}    & $\mathbf{48.31 {\pm} 0.40}$     & $\underline{48.67 {\pm} 0.35}$  & $\underline{40.93 {\pm} 1.02}$  & $\underline{12.59 {\pm} 0.13}$     & $\mathbf{63.08 {\pm} 1.04}$         & $\mathbf{44.68 {\pm} 0.96}$         \\
    \bottomrule
  \end{tabular}
\end{table}

\paragraph{Pairwise significance tests.}
\textbf{(i) Objaverse, matched-data architectural contribution
(no-PV vs.\ PointLLM)}: no-PV outperforms PointLLM on all
six metrics on means (Table~\ref{tab:main-obj}, up to
$+1.48$ GPT-4o judge); Welch's $t$-tests on the four
metrics tabulated above give $p\!\geq\!0.21$ (within
run-to-run noise), consistent with the architectural-level
lift reported in \S\ref{sec:main-results}.
\textbf{(ii) Objaverse, full model with PartVerse-QA (full
vs.\ PointLLM)}: METEOR is significant ($p\!\approx\!0.04$);
GPT-4o ($p\!\approx\!0.09$) and SBERT ($p\!\approx\!0.08$)
are marginal; SimCSE is non-significant ($p\!>\!0.15$).
\textbf{(iii) PartVerse-QA architectural progression
(ours vs.\ +Markers, no MSR)}: this isolates MSR's specific
contribution beyond raw markers. Deterministic
$+0.196$ C2S~Jaccard and $+10.7$ percentage points
C2S~Exact-match (no std applicable) plus
the significant $+2.75$ S2C~GPT-4o judge
($p\!\approx\!0.012$).

\section{Evaluation prompt templates}
\label{app:prompts}
We list the prompt strings used for training supervision
and evaluation. Chat templating matches
PointLLM~\citep{xu2024pointllm}; 3DCoMPaT-GrIn formatting
matches Kestrel~\citep{ahmed2025kestrel}'s release.

\paragraph{Chat template.} All training and inference use the
Vicuna~v1.1~\citep{chiang2023vicuna} chat format with a fixed system
prompt and \texttt{USER}/\texttt{ASSISTANT} roles separated by
a single space, with \texttt{</s>} terminating each
\texttt{ASSISTANT} turn. The \texttt{<point>} placeholder
appears at the start of the first \texttt{USER} turn followed
by a newline, and is replaced at runtime by
\texttt{<point\_start>}, the projected point-cloud token
block, and \texttt{<point\_end>}; subsequent \texttt{USER}
turns in multi-round conversations omit \texttt{<point>}:
\begin{quote}\small\ttfamily
A chat between a curious user and an artificial intelligence
assistant. The assistant gives helpful, detailed, and polite
answers to the user's questions. USER:
<point>$\backslash$n\{instruction\} ASSISTANT:
\{response\}</s>USER: \{follow-up\} ASSISTANT:
\{response\}</s>\dots
\end{quote}
The prompt strings below populate \texttt{\{instruction\}}
for each task.

\paragraph{(a) Objaverse brief-description alignment (Stage~1).}
We use PointLLM's $660$K Cap3D-aligned brief-description
set~\citep{xu2024pointllm}. Each sample draws uniformly from the
released pool of $30$ paraphrases of ``describe this object
briefly''; five representative examples:
\begin{itemize}
  \item ``Offer a clear and concise description of this point
    cloud object.''
  \item ``How would you interpret this 3D point cloud?''
  \item ``What kind of object is illustrated by this
    collection of points?''
  \item ``Convey a summary of the 3D structure represented in
    this point cloud.''
  \item ``Describe the object that this point cloud forms.''
\end{itemize}
The reference response is the Cap3D~\citep{luo2023scalable} brief
caption for that object.

\paragraph{(b) Objaverse complex instructions (Stage~2).}
We use PointLLM's $70$K complex-instruction
set~\citep{xu2024pointllm}, split into
detailed-description ($\sim$15K), single-round
($\sim$40K), and multi-round ($\sim$15K)
GPT-4-generated dialogues conditioned on each object's Cap3D
caption. A representative detailed-description sample:
\begin{quote}\small\ttfamily
USER: <point>$\backslash$nProvide a meticulous explanation
of what these points represent. ASSISTANT: This
three-dimensional model represents a large aircraft with a
dominating white color scheme.\dots</s>
\end{quote}

\paragraph{(c) PartVerse-QA caption-to-slots (training and
evaluation).} Given a part caption, the model returns matching
\partk{k} tokens. The instruction template is:
\begin{quote}\small\ttfamily
<point>$\backslash$nText: ``\{caption\}'' Output only
matching <part\_n> tokens, comma-separated.
\end{quote}
The reference response is comma-separated \partk{k} tokens
without spaces (e.g.\ \texttt{<part\_1>,<part\_4>,<part\_7>}).
The literal string \texttt{<part\_n>} appearing in the
instruction is a generic placeholder in the natural-language
prompt; the model emits actual \partk{k} vocabulary tokens
in its response.

\paragraph{(d) PartVerse-QA slot-to-caption (training and
evaluation).} Given a slot set, the model returns a one-sentence
caption. The instruction template is:
\begin{quote}\small\ttfamily
<point>$\backslash$nDescribe region <part\_$k_1$>,
<part\_$k_2$>, $\dots$, and <part\_$k_n$> in one short
sentence.
\end{quote}
The slot list is comma-separated with ``and'' before the
final token (e.g.\ \texttt{<part\_0>, <part\_5>, and
<part\_13>}).

\paragraph{(e) Objaverse captioning evaluation.} The model is
prompted with the single instruction ``Caption this 3D model
in detail.'', following PointLLM~\citep{xu2024pointllm}.

\paragraph{(f) 3DCoMPaT-GrIn (training and evaluation).}
We fine-tune on the full 3DCoMPaT-GrIn train split
($111{,}514$ samples) released with Kestrel~\citep{ahmed2025kestrel},
which combines three subsets: Part-Aware Point Grounded
Description (PaPGD, $80{,}760$), Direct Segmentation
($8{,}076$), and Reasoning Segmentation ($22{,}678$). Each
subset has its own prompt format; the PaPGD subset uses one
of $30$ paraphrases requesting a part-grounded description
with in-line segmentation masks. Three representative PaPGD
paraphrases:
\begin{itemize}
  \item ``Kindly give me a detailed description of the 3D
    model. Please incorporate interleaved segmentation masks
    for the corresponding components in your answer.''
  \item ``Can you offer a comprehensive analysis of the 3D
    model? Please include interleaved segmentation masks for
    the relevant parts in your response.''
  \item ``Please provide an exhaustive overview of the 3D
    model. Include segmentation masks for each distinct
    component in your answer.''
\end{itemize}
Because 3D-PLOT-LLM is decoder-free, the mask-request phrasing
is treated as text instruction and the model is supervised
only on the language portion of the ground-truth caption.
Evaluation (Table~\ref{tab:compat-papgd}) is on the PaPGD
multi-part split as defined by Kestrel (text track; the grounding-track protocol is in App.~\ref{app:grin-grounded}); fine-tuning
uses the Stage-2 schedule (App.~\ref{app:hparams}) for $3$
epochs.

\paragraph{(g) GPT-4o captioning judge.} Objaverse and
3DCoMPaT-GrIn PaPGD GPT-judge scoring use the prompt
of~\citet{xu2024pointllm} verbatim, with judge model
\texttt{gpt-4o-2024-08-06}:
\begin{quote}\small
``Evaluate a model-generated caption against a human-generated
caption (ground truth) for a 3D model. Identify the aspects
mentioned in the human caption and calculate the percentage
of these aspects correctly mentioned or partially matched in
the model caption. Score from 0 to 100, where each aspect
contributes equally to the score. Consider similar concepts
for partial score. Provide your score (0-100) and a short
justification (less than 15 words) in the format of
`score\#reason'.''
\end{quote}
Cited GPT-4o cells in Tables~\ref{tab:main-obj}
and~\ref{tab:compat-papgd} are blank because GPT-4o snapshots
(and, for the cited Kestrel rows, the judge prompt itself)
differ across papers, so their reported numbers are not
directly comparable to ours under a fixed
\texttt{gpt-4o-2024-08-06} judge.

\section{PartVerse-QA benchmark: construction and statistics}
\label{app:partverse}

\paragraph{Purpose.}
Existing 3D object captioning benchmarks (Objaverse
val~\citep{xu2024pointllm}, PiSA-Bench~\citep{guo2026pisa}) evaluate
whole-object descriptions but contain no part-referring
queries; part-aware benchmarks such as
3DCoMPaT-GrIn~\citep{ahmed2025kestrel,slim20253dcompat++} target segmentation
grounding through pointwise masks, which is orthogonal to the
token-level interface 3D-PLOT-LLM exposes. To evaluate a 3D
MLLM's ability to name parts in its output vocabulary
and to resolve part captions to a finite set of
addressable slots, we construct \textbf{PartVerse-QA}, a
new benchmark we build on top of the part annotations of the
PartVerse mesh dataset~\citep{dong2025one} and align to the
Objaverse point-cloud representation consumed by
PointLLM~\citep{xu2024pointllm} and 3D-PLOT-LLM. The aligned benchmark (query JSONs, alignment and part-to-slot mapping tables, and partition packs) is released under the MIT license of PartVerse at \url{https://huggingface.co/datasets/jintangx/PartVerse-QA}; the evaluation scripts are in the code repository.

\paragraph{Source data.}
All semantic part annotations used by PartVerse-QA come from
the public PartVerse mesh dataset~\citep{dong2025one}; we
introduce no part labels of our own. PartVerse provides meshes
with per-triangle part IDs for a subset of Objaverse-style
objects. Because PointLLM and 3D-PLOT-LLM consume a fixed
$8192$-point sample, we re-project the mesh-level labels onto
the point cloud and then onto our $K{=}16$ region slots. The
pipeline below is deterministic given the PartVerse meshes and
the Objaverse point clouds; no additional human labels are
introduced.

\paragraph{Construction pipeline.}
\begin{enumerate}
  \item \textbf{Point re-projection.} For each object in
    PartVerse, we align its mesh to the 8192-point Objaverse
    sample consumed by PointLLM using an iterative ICP step
    followed by Umeyama rigid alignment, then transfer
    per-triangle part IDs to each of the 8192 points via
    nearest-neighbor on the aligned mesh.
  \item \textbf{Semantic-to-slot mapping.} For each semantic
    part $s$ on an object, we locate the minimum set of region
    slots whose union covers the point support of $s$, and
    record the union-IoU between the slot union and the
    semantic part.
  \item \textbf{IoU filtering (quality control).} We keep a
    (semantic part, slot set) pair only if its union-IoU is
    at least $0.5$. Below this threshold the $K{=}16$ slot
    decomposition is judged too coarse or too ambiguous to
    supervise as a part-grounded description; such pairs are
    discarded. The retained pool has median union-IoU
    $0.713$ (mean $0.725$), well above the $0.5$ floor.
  \item \textbf{Query instantiation.} Each retained
    (semantic part, slot set) pair is instantiated into
    caption-to-slots (C2S) and slot-to-caption
    (S2C) queries (\S\ref{sec:training}); exact prompt
    templates are listed in Appendix~\ref{app:prompts}. C2S
    is scored by set-level Jaccard and Exact-match. S2C is scored by a set
    of standard captioning metrics: Word-F1 (token-level
    overlap), Sentence-BERT (semantic similarity), BLEU-1,
    METEOR, and ROUGE-L, plus a GPT-4o judge for free-form
    semantic agreement; the same metric set is used for
    Objaverse captioning in Table~\ref{tab:main-obj}.
\end{enumerate}

\paragraph{Train / evaluation split.}
The aligned pool contains $10{,}200$ PartVerse objects with
$67{,}521$ raw (semantic part, object) annotations; the
IoU~$\geq 0.5$ filter (Step~3) retains $26{,}065$ unique
(semantic part, slot set) pairs. After an object-level split,
the benchmark yields $77{,}607$ training queries and
$588$ evaluation queries ($392$ C2S, $196$ S2C) on an
$83$-object held-out set. Crucially, the held-out
objects are disjoint not only from our PartVerse-QA training pool
but also from PointLLM's Stage-1 alignment data ($660$K
Cap3D-captioned Objaverse objects) and Stage-2 instruction set
($70$K complex-instruction objects), so the model never sees
any held-out object during pre-training, alignment, or
instruction tuning. Statistical comparisons are reported at
the query level (392 C2S, 196 S2C): each query is an
independent (semantic part, slot set) pair, and the 5-run
inference variance and Welch's $t$-tests
(App.~\ref{app:variance}) operate over query-level scores.

\paragraph{Filtering sensitivity and interface coverage.}
The union-IoU filter of Step~3 is computed from the partition and the mesh annotations alone, with no language model in the loop, and it removes pairs that the $K{=}16$ grid represents poorly. It keeps $38.6\%$ of raw part instances but $73.3\%$ of all part-annotated points: the median removed part spans $186$ points, about a third of one $512$-point slot, against $1{,}447$ points for kept parts. To show what the filter changes, Table~\ref{tab:iou-curve} evaluates the unchanged model on every stratum of ground-truth mapping quality, including evaluation sets rebuilt from the raw annotations for all cases the filter removes ($456$ queries across $[0,0.5)$; parts mapping to zero slots are excluded because no answer key exists), each against a cardinality-matched chance baseline (uniform random slot subsets of the model's predicted cardinality, $1{,}000$ draws per query). Absolute scores decline smoothly as the answer key degrades, while the model-to-chance ratio stays between $2.0\times$ and $2.6\times$ throughout: the threshold changes absolute numbers, not the capability conclusion. The benchmark therefore evaluates addressable coarse grounding on the representable subset, covering about three quarters of the annotated part surface.

\begin{table}[!ht]
  \caption{\textbf{Caption-to-slots as a function of ground-truth mapping quality.} The unchanged model on all strata of the union-IoU between a part and its gold slot set; the three strata at or above $0.5$ make up the $392$-query benchmark. Chance is a cardinality-matched random baseline.}
  \label{tab:iou-curve}
  \centering\small
  \renewcommand{\arraystretch}{0.85}
  \setlength{\tabcolsep}{5pt}
  \begin{tabular}{lcccc}
    \toprule
    Union-IoU stratum & $n$ & Jaccard$\uparrow$ & Chance & Ratio \\
    \midrule
    $\geq 0.7$     & $156$ & $0.552$ & $0.244$ & $2.3\times$ \\
    $[0.6, 0.7)$   & $114$ & $0.471$ & $0.201$ & $2.3\times$ \\
    $[0.5, 0.6)$   & $122$ & $0.330$ & $0.137$ & $2.4\times$ \\
    $[0.4, 0.5)$   & $102$ & $0.318$ & $0.120$ & $2.6\times$ \\
    $[0.2, 0.4)$   & $244$ & $0.176$ & $0.086$ & $2.0\times$ \\
    $[0.0, 0.2)$   & $110$ & $0.147$ & $0.068$ & $2.2\times$ \\
    \bottomrule
  \end{tabular}
\end{table}

\begin{figure}[!ht]
  \centering
  \includegraphics[width=0.70\linewidth]{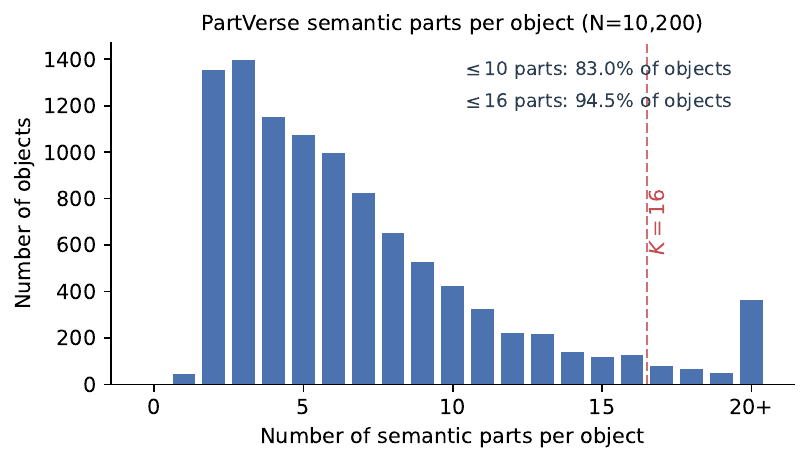}
  \caption{\textbf{Semantic parts per object in our aligned
  PartVerse pool ($N{=}10{,}200$).} Median $6$,
  $p_{95}{=}17$. The dashed line marks our slot budget
  $K{=}16$, which covers $94.5\%$ of objects.}
  \label{fig:partverse-parts-hist}
\end{figure}

\paragraph{Parts per object (pre-filter, $10{,}200$ objects).}
Figure~\ref{fig:partverse-parts-hist} plots the number of
PartVerse semantic-part annotations per aligned object,
before any IoU filtering. The distribution is
concentrated in the low single digits: median $6$,
$p_{90}{=}13$, $p_{95}{=}17$; $83\%$ of objects carry at
most $10$ parts and $94.5\%$ carry at most $16$. This grounds
our $K{=}16$ default. A smaller $K$ (e.g., $K{=}8$) would
leave a substantial fraction of objects' parts overflowing
the slot budget and force semantic parts to share slots,
degrading the part-token interface (App.~\ref{app:sensitivity}: C2S point-IoU $0.460 \to 0.406$ at $K{=}8$); a larger $K$ (e.g.,
$K{=}32$) would double the \partk{k} vocabulary and dilute
the average per-token training signal, with diminishing
coverage gains beyond the $95$th percentile and finer slots
that fragment beyond the training-free partition's effective
resolution against semantic parts.
A direct comparison across $K$ is confounded by the filter, which selects different (semantic part, slot set) pairs at different $K$ (the $K{=}8$ held-out set built by the same pipeline shares $69$ objects with the $K{=}16$ set). App.~\ref{app:sensitivity} therefore reports a full $K{=}8$ retrain scored with point-IoU, which is defined in point space and comparable across $K$; the architectural argument itself is $K$-agnostic (any spatially coherent partition with the two properties of \S\ref{sec:bcp} would serve).

\begin{figure}[!ht]
  \centering
  \includegraphics[width=0.70\linewidth]{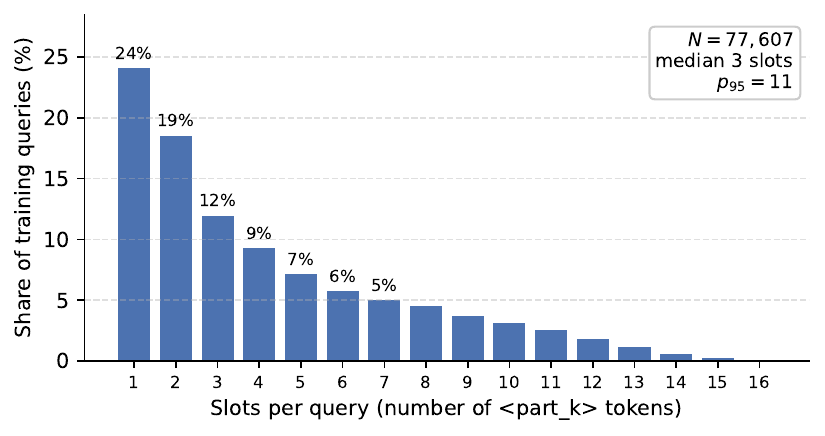}
  \caption{\textbf{Slots per query across $77{,}607$ training
  queries.} The benchmark spans single-slot fine-grained parts
  (24\%) through near-whole-object regions ($11$ slots at
  $p_{95}$, $16$ slots at the tail). Eval splits
  ($392$ C2S + $196$ S2C) follow the same distribution.}
  \label{fig:partverse-slotset-hist}
\end{figure}

\paragraph{Slots per query (post-filter, $77{,}607$ queries).}
Figure~\ref{fig:partverse-slotset-hist} plots the slot-set
size per training query, after the IoU~$\geq 0.5$
filter and query expansion. The benchmark covers a wide
granularity range: 24\% of queries target a single slot
(fine-grained part references), the median query has $3$
slots, and the 95th percentile reaches $11$ slots
(near-whole-object regions). C2S and S2C share the same
cardinality distribution by construction.

\paragraph{Faithfulness to PartVerse annotations.}
We do not modify PartVerse's raw mesh annotations; our
additions (re-projection onto the 8192-point Objaverse sample,
slot-union mapping onto the $K{=}16$ grid, the IoU~$\geq 0.5$
filter, and query-template instantiation) operate
deterministically on the public PartVerse meshes and Objaverse
point clouds.

\paragraph{Cross-method evaluation.}
This subsection deepens the structural argument summarized in
\S\ref{sec:main-results} (PartVerse-QA paragraph) and details
why prior part-aware 3D MLLMs are not retrofitted onto either direction of
the benchmark. Prior part-aware 3D MLLMs treat part references
as decoder outputs only: segmentation-based methods
(Kestrel~\citep{ahmed2025kestrel},
PARIS3D~\citep{kareem2024paris3d},
SegPoint~\citep{he2024segpoint}) emit point masks via a
\texttt{[SEG]} token decoded by an external head; box-based
Part-X-MLLM~\citep{wang2025part} emits autoregressive box
tokens addressing coordinate bins; none read or write
\partk{k} tokens, and none accept a region encoding (mask,
box, or token set) as model input.
\textbf{The S2C direction (slot set as input)} is therefore
not retrofittable: there is no input pathway in any prior
method through which a region specification can reach the
LLM, and any natural-language translation of the slot set
would leak the target into the prompt and trivialize the
task.
\textbf{The C2S direction (slot set as output)} is in
principle retrofittable by projecting a method's predicted
masks or boxes onto our $K{=}16$ slot grid. Kestrel and Part-X-MLLM have no public checkpoints, and PARIS3D and SegPoint take rendered views or scene-level inputs that would need per-method adaptation, an engineering layer between the model and the metric; App.~\ref{app:find3d} projects the open-vocabulary part segmenter Find3D instead. Since the projection adds its own error, App.~\ref{app:find3d} also reports Jaccard separately for parts that the slot grid covers well and poorly. App.~\ref{app:mesh-miou} scores our own variants against mesh-level masks without any slot projection. The slot-mapping tables are part of the released benchmark, so future work can build unified cross-method evaluations on top.

\section{Model-level partition-agnostic mesh mIoU}
\label{app:mesh-miou}

This appendix complements
Appendix~\ref{app:partverse}, which establishes that the
$K{=}16$ partition's median union-IoU against PartVerse
mesh-level semantic parts is $0.713$. We extend that
partition-level statistic to a model-level mIoU computed
against mesh-derived per-point masks rather than slot-set
ground truth, providing a partition-agnostic view of the
architectural progression in
Table~\ref{tab:component-ablation}.

\paragraph{Method.}
For each held-out C2S query we lift the model's predicted
\partk{k} set to the $8192$-point Objaverse cloud via the
PointBERT-Group rule used at training, and build the
mesh-level ground-truth mask by ICP$+$Umeyama alignment of
the cached PartVerse mesh-sampled cloud onto the same point
cloud; alignments are pre-computed at PartVerse-QA
construction (Appendix~\ref{app:partverse}) and reused here.
$390$ of $392$ queries pass; two are dropped at fewer than
$50$ mesh points after alignment.

\begin{table}[!ht]
  \caption{\textbf{Mesh mIoU on PartVerse-QA C2S
  ($N{=}390$).} Set Jaccard from
  Tables~\ref{tab:main-partverse} and~\ref{tab:component-ablation}
  for reference. Mesh GT comes from PartVerse mesh annotations
  ICP-aligned to the $8192$-point cloud, with no slot-set
  reprojection. Rows span the variants of
  Tables~\ref{tab:component-ablation} and~\ref{tab:msr-branch}.}
  \label{tab:mesh-miou}
  \centering
  \footnotesize
  \begin{tabular}{lccc}
    \toprule
    Variant
      & Set Jaccard$\uparrow$
      & Mesh mIoU mean$\uparrow$
      & Mesh mIoU median$\uparrow$ \\
    \midrule
    Vocab-only (no markers)
      & $0.413$ & $0.374$ & $0.374$ \\
    + Markers (no MSR)
      & $0.263$ & $0.257$ & $0.192$ \\
    + MSR stats only ($\alpha_{\text{adj}}{=}0$)
      & $0.284$ & $0.287$ & $0.239$ \\
    + MSR adjacency only ($\alpha_{s}{=}0$)
      & $0.305$ & $0.303$ & $0.254$ \\
    \textbf{+ MSR full (ours)}
      & $\mathbf{0.459}$ & $\mathbf{0.399}$ & $\mathbf{0.426}$ \\
    \midrule
    Gold slot set (slot-set ceiling)
      & $1.000$ & $0.688$ & $0.669$ \\
    Partition oracle ($K{=}16$ ceiling)
      & --- & $0.711$ & $0.705$ \\
    \bottomrule
  \end{tabular}
\end{table}

\paragraph{Findings.}
Three observations across Table~\ref{tab:mesh-miou}.
(i) The architectural ranking is preserved under the
partition-agnostic GT (MSR full $>$ Vocab-only $>$ MSR
adjacency-only $>$ MSR stats-only $>$ Markers without MSR),
including the markers-no-MSR regression below vocab-only.
(ii) The partition oracle's mean mesh mIoU $0.711$ matches
the median union-IoU $0.713$ from
Appendix~\ref{app:partverse}, confirming pipeline consistency.
(iii) Dense-decoder methods (PARIS3D, SegPoint, Kestrel) are
not directly comparable on this metric: their structural
ceiling exceeds any $K{=}16$-based ceiling (e.g., the $0.711$
partition oracle here), the explicit design tradeoff of this
work (\S\ref{sec:related}, Appendix~\ref{app:partverse}
``Cross-method evaluation'').

\section{MSR branch ablation: stats vs.\ adjacency}
\label{app:msr-branch}

MSR has two residual branches (Eqs.~\ref{eq:stats}--\ref{eq:graph}):
a stats branch driven by per-region statistics
$s_k\!\in\!\R^7$ via $\text{MLP}(s_k)$, and an adjacency branch
driven by message passing over $\mathcal{N}(k)$ via
$\text{MP}(\tilde{\mathbf{m}}_k;\mathcal{N}(k))$.
Table~\ref{tab:msr-branch} ablates each branch in isolation
against the full two-branch MSR.

\begin{table}[!ht]
  \caption{\textbf{MSR branch ablation.} Stats-only and
  adjacency-only each achieve a substantial fraction of the
  full MSR's gains, but neither alone matches the full
  configuration on the part-grounded objective. The two
  branches are individually informative and complementary:
  combining them lifts PartVerse-QA Jaccard by $+0.154$ over
  adjacency-only and $+0.175$ over stats-only.}
  \label{tab:msr-branch}
  \centering\small
  \renewcommand{\arraystretch}{0.85}
  \setlength{\tabcolsep}{4pt}
  \begin{tabular}{lcccc}
    \toprule
    & \multicolumn{2}{c}{Objaverse} & \multicolumn{2}{c}{PartVerse-QA} \\
    \cmidrule(lr){2-3} \cmidrule(lr){4-5}
    MSR branch active            & SBERT$\uparrow$     & GPT-4o$\uparrow$    & C2S Jac$\uparrow$  & S2C GPT-4o$\uparrow$ \\
    \midrule
    Stats only ($\alpha_{\text{adj}}{=}0$)            & $\underline{47.88}$ & $\underline{40.26}$ & $0.284$            & $\underline{43.93}$ \\
    Adjacency only ($\alpha_{\text{s}}{=}0$)          & $47.31$             & $39.21$             & $\underline{0.305}$& $43.69$             \\
    \textbf{Both (ours)}                              & $\mathbf{48.31}$    & $\mathbf{40.93}$    & $\mathbf{0.459}$   & $\mathbf{44.68}$    \\
    \bottomrule
  \end{tabular}
\end{table}

\paragraph{Branches are complementary, not redundant.}
Each isolated branch reaches a substantial but incomplete
fraction of the full-MSR PartVerse-QA Jaccard, and combining
them exceeds the sum of individual-branch improvements over
the no-MSR baseline ($0.263$, Table~\ref{tab:component-ablation}):
the combined gain is $+0.196$, versus $+0.021$ (stats-only)
and $+0.042$ (adjacency-only) summing to $+0.063$. The
asymmetry across metrics is interpretable: stats-only leads on
Objaverse captioning (per-region geometry signals whole-object
structure), and adjacency-only leads on PartVerse-QA Jaccard
(inter-region adjacency binds part references to slot sets),
with the two branches complementing in the full model.

\section{Probing token-region binding}
\label{app:permutation}

To check whether \partk{k} functions as a real addressing
handle rather than a cosmetic label, we permute the
\partk{k} identifiers by a fixed random $\sigma$ at
inference while keeping the partition, markers, and patches
unchanged (no retraining). If the binding is real, the C2S
predictions should shift off the original label space and
Jaccard against the original GT should collapse.
Table~\ref{tab:permutation} confirms this: Jaccard drops
from $0.459$ to $0.209$ and Exact-match from $13.78\%$ to
$1.28\%$, well below single-slot chance ($1/16{=}6.25\%$),
which is evidence that the LLM has learned a genuine
token-region binding.

\begin{table}[!ht]
  \caption{\textbf{Label-permutation probe on PartVerse-QA
  C2S ($392$ held-out queries).} Partition, markers, and
  patches unchanged; \partk{k} labels permuted at inference.}
  \label{tab:permutation}
  \centering\small
  \renewcommand{\arraystretch}{0.85}
  \setlength{\tabcolsep}{4pt}
  \begin{tabular}{lcc}
    \toprule
    Setting & Jaccard$\uparrow$ & Exact-match$\uparrow$ \\
    \midrule
    Baseline (ours) & $\mathbf{0.459}$ & $\mathbf{13.78\%}$ \\
    Label permutation ($\sigma$ random) & $0.209$ & $1.28\%$ \\
    \bottomrule
  \end{tabular}
\end{table}

\section{Design space of refinement module}
\label{app:refinement-alt}

This appendix expands the single-line mention in
\S\ref{sec:ablation}: we ablate where the
stats~$+$~adjacency conditioning module is applied: to the
per-region marker (our MSR), to the vocabulary token
\partk{k} (``vocab refinement''), or to both. The only
varying axis across rows is the channel that receives the
conditioning.

\begin{table}[!ht]
  \caption{\textbf{Design space of refinement module:
  where to apply structural conditioning.} Rows share the
  ``$+$ Markers (no MSR)'' assembly of
  Table~\ref{tab:main-partverse} and differ only in the
  conditioning target: nowhere; the $4096$-d \partk{k}
  vocabulary embedding; the $384$-d marker $\mathbf{m}_k$
  (ours, MSR); or both. C2S Jaccard is deterministic; S2C
  GPT-4o is the $5$-run mean on the PartVerse-QA
  $196$-query held-out split (per-row $\pm$std and Welch's
  $t$-tests in prose below).}
  \label{tab:refinement-alt}
  \centering\small
  \renewcommand{\arraystretch}{0.85}
  \setlength{\tabcolsep}{4pt}
  \begin{tabular}{lccc}
    \toprule
    Refinement location & C2S Jaccard$\uparrow$ & S2C GPT-4o$\uparrow$ & Params$\downarrow$ \\
    \midrule
    No refinement                          & $0.263$              & $41.93$              & $\mathbf{0}$       \\
    Vocab refinement                       & $0.278$              & $42.58$              & $3.2$M             \\
    Marker $+$ vocab refinement            & $\underline{0.442}$  & $\underline{43.39}$  & $3.9$M             \\
    \textbf{Marker refinement (ours, MSR)} & $\mathbf{0.459}$     & $\mathbf{44.68}$     & $\underline{0.7}$M \\
    \bottomrule
  \end{tabular}
\end{table}

\paragraph{Findings.} Three observations are consistent
across metrics in
Table~\ref{tab:refinement-alt}:

\textbf{(i) Marker refinement dominates.}
Applying the conditioning module to the marker (MSR) lifts
C2S Jaccard from $0.263$ to $0.459$ ($+0.196$); applying the
same module to the vocab token instead (vocab
refinement) lifts Jaccard only to $0.278$ ($+0.015$). The
asymmetry is $\sim\!13\!\times$ on Jaccard. On S2C GPT-4o,
marker refinement also leads: $+2.75$ over no refinement and
$+2.10$ over vocab refinement.

\textbf{(ii) Marker location is more parameter-efficient.}
MSR achieves the best result with $0.7$M extra parameters;
the vocab-side variant requires $3.2$M
($\sim\!4.5\!\times$ more) because each \partk{k} embedding
is $4096$-d vs the marker's $384$-d. The same conditioning
loss thus provides denser per-parameter gradient on the
marker.

\textbf{(iii) Combining is non-additive.} Refining both
locations reaches $0.442$, $-0.017$ below MSR alone,
suggesting the two channels encode mutually interfering
inductive biases when stacked rather than additive structural
signal. We therefore use single-location MSR.

\section{PartVerse-QA data scaling}
\label{app:data-scaling}
\begin{figure}[!ht]
  \centering
  \includegraphics[width=0.82\linewidth]{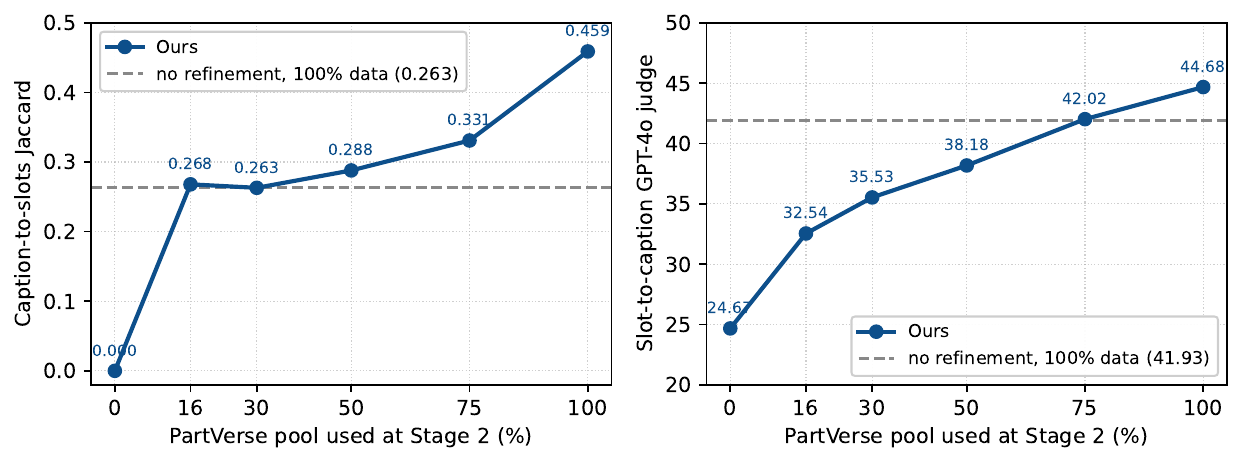}
  \caption{\textbf{PartVerse-QA supervision scaling.}
  Caption-to-slots Jaccard (left) and slot-to-caption GPT-4o
  judge (right) on the held-out PartVerse-QA split as a function
  of the fraction of the $77{,}607$-pair pool used at
  Stage~2 (six points: $0/16/30/50/75/100\%$). The dashed
  horizontal references mark the no-refinement comparator at
  $100\%$ data; ours meets the no-refinement Jaccard reference
  using only $16\%$ of the pool, and the no-refinement
  slot-to-caption GPT-4o judge by $75\%$.}
  \label{fig:data-scaling}
\end{figure}

Figure~\ref{fig:data-scaling} and
Table~\ref{tab:data-scaling} report 3D-PLOT-LLM trained on
$0/16/30/50/75/100\%$ of the $77{,}607$ PartVerse-QA pairs at
Stage~2, with the Markers baseline of
Table~\ref{tab:main-partverse} as the no-refinement reference.
Three observations stand out:

(i)~\textbf{Sample efficiency.} Ours with just $16\%$
PartVerse-QA data already matches the no-refinement Markers
baseline trained on the full pool (caption-to-slots Jaccard
$0.268$ vs.\ $0.263$, Table~\ref{tab:main-partverse}): a
small slice of part-grounded supervision is enough to
reach a baseline-equivalent grounding signal.

(ii)~\textbf{Steep high-data scaling.} Ours rises to
$0.459$ Jaccard at $100\%$, with the steepest segment in
$75\!\to\!100\%$ ($+0.128$ Jaccard, $+2.66$ slot-to-caption
GPT-4o judge), and ultimately a $+0.196$ Jaccard margin over
the no-refinement Markers baseline
(Table~\ref{tab:main-partverse}) at full data.

(iii)~\textbf{Objaverse is flat.} Whole-object Objaverse
captioning varies little across fractions
($47.6$--$48.3$ SBERT, $38.8$--$40.9$ GPT-4o), consistent
with Stage-2 captioning loss being saturated by the shared
$70$K complex-instruction pool that every variant trains on.

\begin{table}[!ht]
  \caption{\textbf{PartVerse-QA supervision scaling.} Ours
  with only $16\%$ of the PartVerse-QA pool already matches
  the no-refinement baseline (Markers row in
  Table~\ref{tab:main-partverse}, Jaccard $0.263$) trained on
  the full pool. Fractions: share of the $77{,}607$-pair
  PartVerse-QA pool used at Stage~2.}
  \label{tab:data-scaling}
  \centering\small
  \renewcommand{\arraystretch}{0.85}
  \setlength{\tabcolsep}{4pt}
  \begin{tabular}{lcccc}
    \toprule
    Fraction & PV-Jac$\uparrow$ & Obj-SBERT$\uparrow$ & Obj-GPT-4o$\uparrow$ & S2C-GPT-4o$\uparrow$ \\
    \midrule
    $0\%$   & $0.000$              & $47.94$              & $40.56$              & $24.67$              \\
    $16\%$  & $0.268$              & $47.65$              & $\underline{40.77}$  & $32.54$              \\
    $30\%$  & $0.263$              & $47.74$              & $38.83$              & $35.53$              \\
    $50\%$  & $0.288$              & $\underline{48.00}$  & $40.75$              & $38.18$              \\
    $75\%$  & $\underline{0.331}$  & $47.85$              & $40.69$              & $\underline{42.02}$  \\
    $100\%$ & $\mathbf{0.459}$     & $\mathbf{48.31}$     & $\mathbf{40.93}$     & $\mathbf{44.68}$     \\
    \bottomrule
  \end{tabular}
\end{table}

\section{Sensitivity to the slot count and to the partition}
\label{app:sensitivity}

\paragraph{Slot count.}
We trained a complete $K{=}8$ counterpart of the full model (same recipe, both stages) and evaluated it under the identical protocol on the $K{=}8$ version of PartVerse-QA built by the same pipeline ($258$ C2S and $129$ S2C held-out queries on the $69$ objects shared with the $K{=}16$ split). Across $K$ we compare point-IoU for caption-to-slots, which is defined in point space and comparable across slot granularities, unlike slot-set Jaccard. Table~\ref{tab:k-sensitivity} shows three consistent effects: whole-object captioning at $K{=}8$ falls back to the PointLLM level ($47.05 \pm 0.21$ vs $47.66 \pm 0.58$), part description is unchanged (S2C SBERT $62.30$ vs $62.04$), and part grounding degrades moderately (point-IoU $0.460 \to 0.406$), since eight slots are too coarse to isolate typical PartVerse parts (the mean part occupies $4.5$ of $16$ slots). The dependence on $K$ is smooth rather than knife-edge; $K{=}16$ is set by the $512$-patch budget and was not tuned on benchmark scores.

\begin{table}[!ht]
  \caption{\textbf{Slot-count sensitivity.} Full retrain at $K{=}8$ vs the $K{=}16$ model, same recipe and protocol; $5$-run mean$\,\pm\,$std where applicable.}
  \label{tab:k-sensitivity}
  \centering\small
  \renewcommand{\arraystretch}{0.85}
  \setlength{\tabcolsep}{5pt}
  \begin{tabular}{lcc}
    \toprule
    & $K{=}8$ & $K{=}16$ (paper) \\
    \midrule
    Objaverse SBERT                     & $47.05 \pm 0.21$ & $48.31 \pm 0.40$ \\
    PartVerse-QA S2C SBERT              & $62.30 \pm 0.92$ & $62.04 \pm 1.21$ \\
    PartVerse-QA C2S point-IoU          & $0.406$          & $0.460$ \\
    \bottomrule
  \end{tabular}
\end{table}

\paragraph{Partition hyperparameters and partition family.}
The partition of App.~\ref{app:bcp} has two graph hyperparameters, the $k$NN degree ($k{=}12$) and the long-edge pruning percentile ($75$th). We regenerated the partition at several settings with the original pipeline and evaluated the unchanged $K{=}16$ model with each partition swapped in at inference (no retraining). The regenerated default reproduces the native partition on $96.6\%$ of patch labels, whereas the variants genuinely alter it ($55$ to $58\%$ agreement). Across a $4\times$ range of $k$ and the whole pruning range the spread is $0.8$ SBERT, comparable to $5$-run noise, and no setting falls below the PointLLM baseline beyond noise. Absolute levels in Table~\ref{tab:partition-sweep} are not directly comparable with Table~\ref{tab:main-obj}, because the swap pathway rebuilds slot indices, statistics, and adjacency at load time; rows within Table~\ref{tab:partition-sweep} share this pathway and are comparable with one another. Swapping to a k-means partition of the patch coordinates (a different family, purely spatial) gives $47.74 \pm 0.26$ on captioning and, with the ground truth remapped to that partition by the identical mapping pipeline, $0.446$ vs $0.459$ Jaccard on caption-to-slots; the two answer keys are equally faithful (mean gold union-IoU $0.677$ k-means vs $0.686$ native). The graph hyperparameters are therefore not load-bearing, and the feature term of the merge cost is not either; we keep the $k$NN construction for determinism and local coherence.

\begin{table}[!ht]
  \caption{\textbf{Partition sensitivity at inference.} The unchanged $K{=}16$ model with regenerated partitions swapped in; Objaverse SBERT, $5$-run mean$\,\pm\,$std, all rows through the same swap pathway.}
  \label{tab:partition-sweep}
  \centering\small
  \renewcommand{\arraystretch}{0.85}
  \setlength{\tabcolsep}{5pt}
  \begin{tabular}{lc}
    \toprule
    Partition at inference & SBERT$\uparrow$ \\
    \midrule
    $k$NN $k{=}12$, prune at $75$th pct (default, regenerated) & $47.55 \pm 0.41$ \\
    $k$NN $k{=}6$ (halved)                                    & $48.07 \pm 0.33$ \\
    $k$NN $k{=}24$ (doubled)                                  & $48.24 \pm 0.25$ \\
    Prune at $50$th pct (tighter)                             & $47.71 \pm 0.26$ \\
    No pruning                                                & $47.42 \pm 0.44$ \\
    k-means ($16$) on patch coordinates (different family)    & $47.74 \pm 0.26$ \\
    \bottomrule
  \end{tabular}
\end{table}

\section{Projected open-vocabulary baseline on caption-to-slots}
\label{app:find3d}

This appendix gives the protocol behind the Find3D comparison in \S\ref{sec:main-results}. Among part-aware 3D MLLMs, Kestrel and Part-X-MLLM have no public checkpoints at the time of writing, and PARIS3D and SegPoint take rendered views or scene-level inputs that would need per-method adaptation. The strongest reproducible representative of the open-vocabulary text-to-part-mask family that is point-cloud native and takes free-text queries is Find3D~\citep{ma2025find}; it is trained on Objaverse, so it is in-domain for the held-out set. For each of the $392$ C2S queries, Find3D receives the caption as its text query; its per-point similarity is converted to a query probability (softmax over the query logit and an unlabeled logit) and mean-pooled per slot. A slot set is obtained by (a) a probability threshold swept over $\{0.3, 0.4, 0.5, 0.6, 0.7\}$, reporting the best ($0.5$), and (b) an oracle-cardinality rule that takes the top-$m$ slots with $m$ equal to the gold cardinality, a generous upper bound. Scoring uses the identical slot-to-point mapping and metrics as our model (gold-vs-gold IoU $1.0$ under the shared mapping). Find3D's text encoder truncates long captions at $64$ tokens, a property of that baseline. Table~\ref{tab:find3d} reports the results overall and per mapping-quality stratum.

\begin{table}[!ht]
  \caption{\textbf{Projected open-vocabulary baseline on PartVerse-QA C2S ($392$ queries).} Find3D masks projected onto the $K{=}16$ slot grid; strata by gold union-IoU as in Table~\ref{tab:iou-curve}.}
  \label{tab:find3d}
  \centering\small
  \renewcommand{\arraystretch}{0.85}
  \setlength{\tabcolsep}{4pt}
  \begin{tabular}{lccccc}
    \toprule
    & \multicolumn{2}{c}{All $392$ queries} & \multicolumn{3}{c}{Jaccard by gold union-IoU} \\
    \cmidrule(lr){2-3}\cmidrule(lr){4-6}
    & Jaccard$\uparrow$ & Point-IoU$\uparrow$ & $[0.5,0.6)$ & $[0.6,0.7)$ & $\geq 0.7$ \\
    \midrule
    Find3D, best threshold ($0.5$)       & $0.304$          & $0.311$          & $0.227$          & $0.315$          & $0.356$ \\
    Find3D, oracle cardinality           & $0.294$          & $0.294$          & $0.169$          & $0.300$          & $0.388$ \\
    \textbf{3D-PLOT-LLM (ours)}          & $\mathbf{0.459}$ & $\mathbf{0.460}$ & $\mathbf{0.330}$ & $\mathbf{0.471}$ & $\mathbf{0.552}$ \\
    \bottomrule
  \end{tabular}
\end{table}

\section{3DCoMPaT-GrIn grounding track}
\label{app:grin-grounded}

This appendix gives the protocol behind the grounding-track paragraph of \S\ref{sec:main-results}. Ground-truth masks come from the official fine-grained per-point part labels of the 3DCoMPaT++ validation and test splits, transferred by nearest neighbor to the $8{,}192$-point clouds our models take as input; labels are transferred for all $7{,}443$ GrIn validation objects, with a mean transfer distance of $0.024$. The queries are the part names of the GrIn validation annotations: $677$ single-part, $2{,}000$ multiple-parts, and $2{,}002$ embodied queries. One query each in the multiple-parts and embodied splits is dropped because its ground-truth mask is empty. The PartVerse-QA-trained $K{=}16$ model answers each query zero-shot with greedy decoding. The text-track model of Table~\ref{tab:compat-papgd} is not used here, because it is trained without PartVerse-QA and never learns to answer with slots. The instruction is:
\begin{quote}\small\ttfamily
<point>$\backslash$nThe following text describes ONE component of this 3D object:$\backslash$n``\{part\}''$\backslash$nWhich pseudo-part tokens (<part\_0> through <part\_15>) best correspond to this component? Answer with comma-separated tokens only (e.g. ``<part\_0>,<part\_15>''), with no other text.
\end{quote}
The predicted slot set is turned into a point mask by taking the union of its patches' points, the rule used in App.~\ref{app:mesh-miou}, and scored against the ground-truth mask by mIoU, acc@25, and acc@50, where acc@25 and acc@50 are the fractions of queries with IoU of at least $0.25$ and $0.50$. The oracle ceiling is the mean over queries of the best IoU that any slot set on the $K{=}16$ grid can reach; it measures the limit set by the grid, not the model. The model emits a slot set for every query and reaches $49$ to $50\%$ of the ceiling on all three splits (Table~\ref{tab:grin-grounded}).

\begin{table}[!ht]
  \caption{\textbf{Zero-shot grounding on 3DCoMPaT-GrIn.} Slot-induced point masks of the PartVerse-QA-trained model against per-point fine part labels; the oracle ceiling is the best mIoU any slot set on the $K{=}16$ grid can reach.}
  \label{tab:grin-grounded}
  \centering\small
  \renewcommand{\arraystretch}{0.85}
  \setlength{\tabcolsep}{5pt}
  \begin{tabular}{lccccc}
    \toprule
    Split & $n$ & mIoU$\uparrow$ & acc@25$\uparrow$ & acc@50$\uparrow$ & Oracle ceiling \\
    \midrule
    single\_part     & $677$  & $0.221$ & $0.332$ & $0.158$ & $0.448$ \\
    multiple\_parts  & $1{,}999$ & $0.220$ & $0.326$ & $0.147$ & $0.441$ \\
    embodied         & $2{,}001$ & $0.242$ & $0.379$ & $0.176$ & $0.488$ \\
    \bottomrule
  \end{tabular}
\end{table}

\section{Qualitative examples and failure modes}
\label{app:qualitative}

This appendix supplements \S\ref{sec:qualitative} and
Figure~\ref{fig:qualitative} with untruncated outputs across
all three benchmarks, plus a discussion of representative
failure modes. App.~\ref{app:qual:partverse}
expands the PartVerse-QA bidirectional dialogues across the
slot-cardinality range; App.~\ref{app:qual:grin} reproduces the
full grounded descriptions of Figure~\ref{fig:qualitative}(d-f)
and adds three more 3DCoMPaT-GrIn cases at full length;
App.~\ref{app:qual:obja} provides Objaverse whole-object
side-by-side captions against PointLLM and ShapeLLM, the
qualitative companion to Table~\ref{tab:main-obj};
App.~\ref{app:qual:limits} delineates three failure modes that
characterize where the $K{=}16$ vocabulary-level addressing
interface breaks down.

\subsection{PartVerse-QA: extended bidirectional dialogues}
\label{app:qual:partverse}

\begin{figure}[!htbp]
  \centering
  \includegraphics[width=0.95\linewidth]{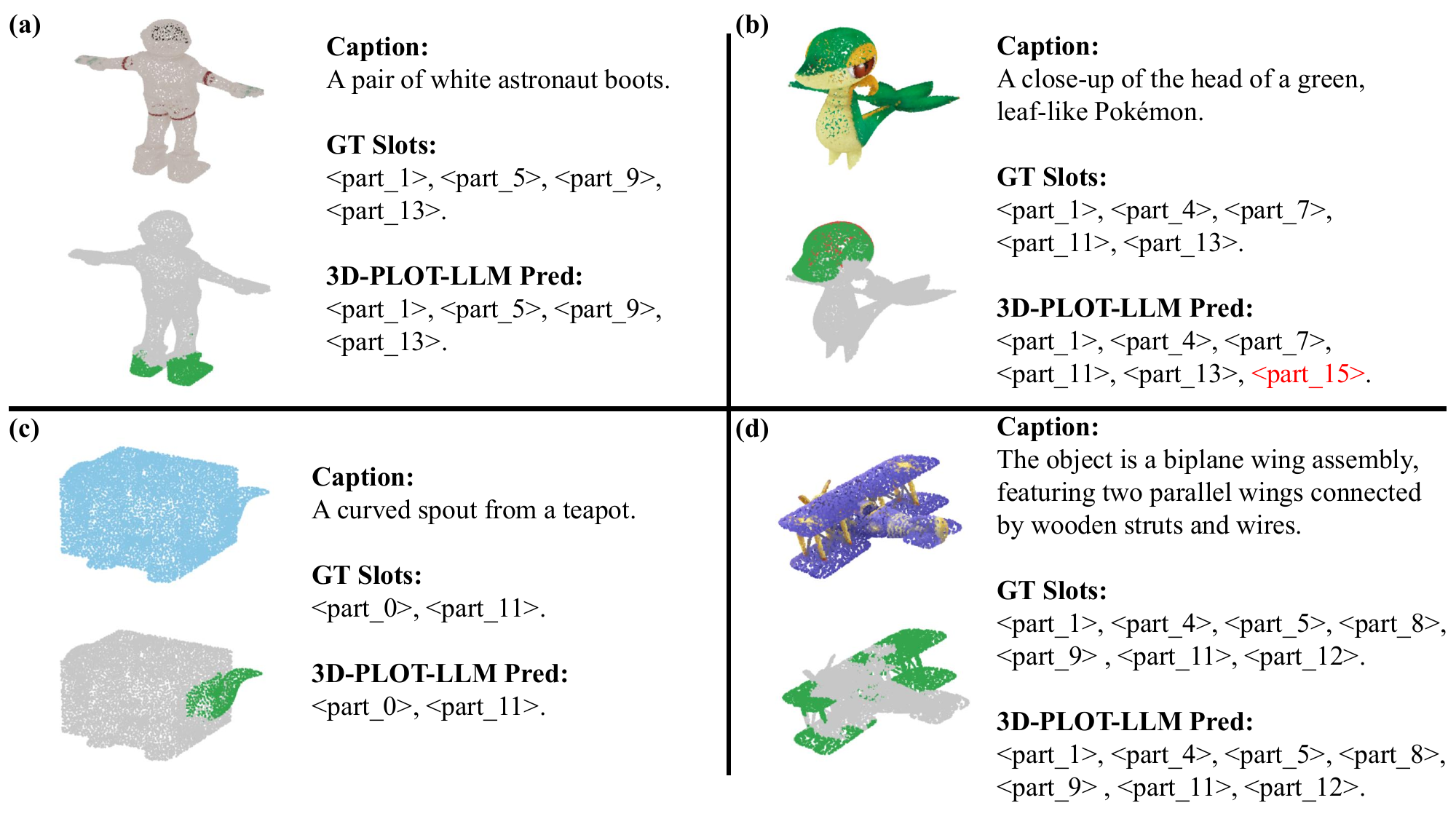}
  \caption{\textbf{PartVerse-QA C2S dialogues (extended).} Each panel
  shows the RGB input (top) and the $K{=}16$ partition with target
  slots highlighted in green (bottom; in panel (b), the additional
  slot the model emits is shown in red), alongside the caption and
  predicted slot set. Panels (a),
  (c), and (d) are exact set matches at $4$, $2$, and $7$ slots
  respectively; panel (b) is a near-miss at $5$ slots where the extra
  red slot \partk{15} is itself inside the head region. Panel (d)
  also shows that token-level set agreement is attainable even when
  the green partition highlight does not visually cover the full upstream
  PartVerse part annotation: token-Jaccard measures slot-token
  agreement, which is independent of how the partition aligns with
  the PartVerse annotation.}
  \label{fig:appx-partverse-c2s}
\end{figure}

Figure~\ref{fig:appx-partverse-c2s} extends the
Caption-to-Slots (C2S) interface beyond the single
Jaccard $=1.00$ panel in the main figure. Four panels span
slot cardinality $2$ to $7$ out of $K{=}16$: two clean exact
matches (a teapot spout at $2$ slots, a pair of astronaut
boots at $4$ slots), one near-miss on a Pok\'emon-style head
($5$ slots, Jaccard $0.83$) where the model emits one extra
slot that also lies inside the head region, and a biplane
wing ($7$ slots) at token-level exact match (Jaccard $=1.00$)
where the green partition highlight does not cover the full
PartVerse ``wing assembly'' annotation, illustrating that
token-Jaccard measures slot-token agreement and is
independent of how the partition aligns with the PartVerse
annotation.

\begin{figure}[!htbp]
  \centering
  \includegraphics[width=0.95\linewidth]{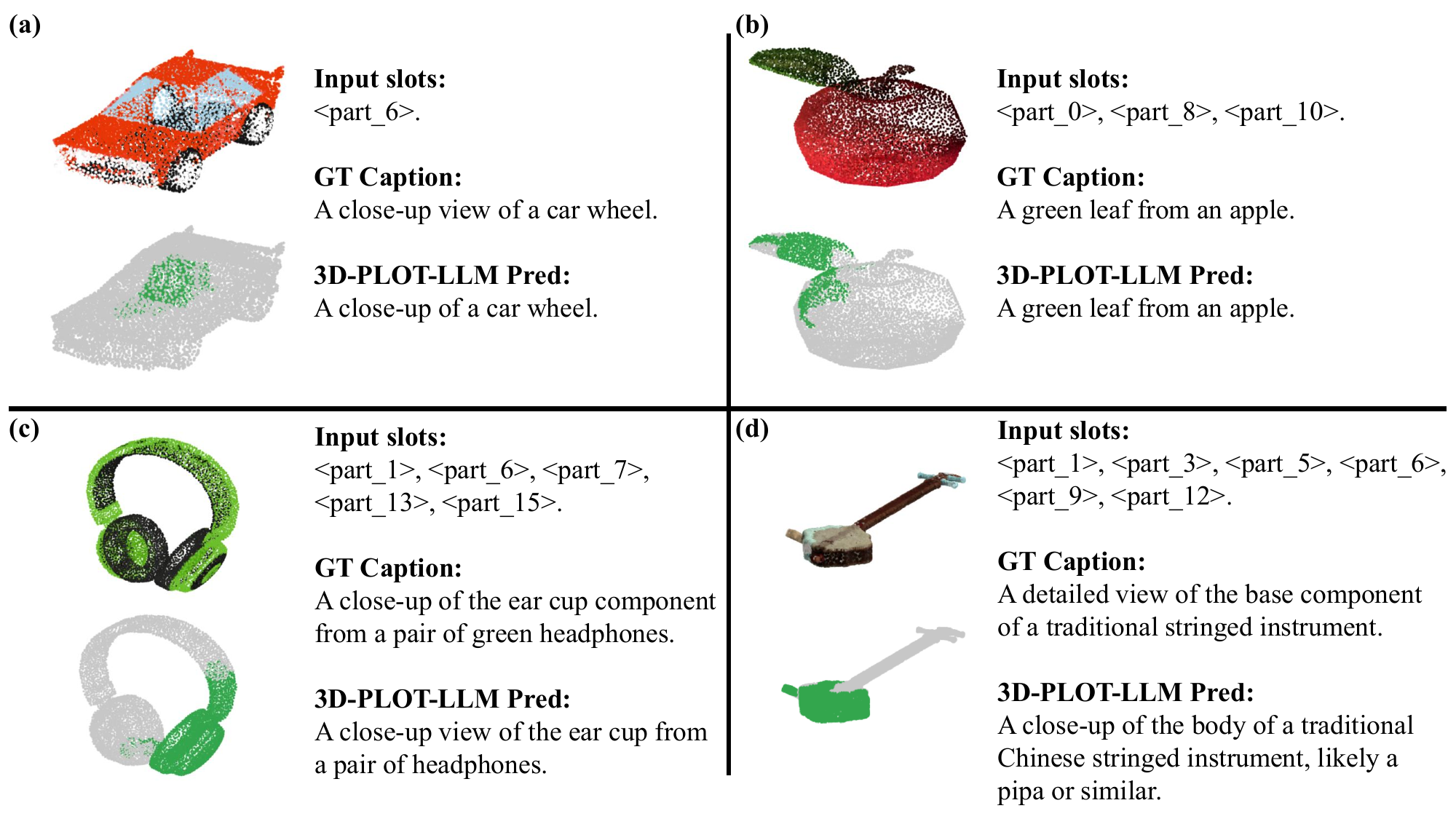}
  \caption{\textbf{PartVerse-QA S2C dialogues (extended).} Each panel
  shows the RGB input (top) and the $K{=}16$ partition with input
  slots highlighted (bottom, in green), alongside the GT caption
  and 3D-PLOT-LLM prediction.
  Panel (a) is a single-word omission ($1$ slot, ``view'' dropped);
  panel (b) is an exact $3$-slot match; panel (c) drops a color
  adjective and a part-label modifier (``green''/``component''); panel
  (d) shows a case where the prediction is more descriptive than
  the GT (``body'' as a more anatomically precise term, plus the
  cultural qualifier ``Chinese'' and a hedged species guess ``likely
  a pipa or similar''), illustrating that Word-F1 penalizes
  additional specificity even when the additions are plausible.}
  \label{fig:appx-partverse-s2c}
\end{figure}

Figure~\ref{fig:appx-partverse-s2c} extends the
Slots-to-Caption (S2C) direction in the same way, with slot
cardinality $1$ to $6$ out of $K{=}16$. Panels include a
near-perfect single-slot car wheel (``\dots view of a car
wheel'' $\to$ ``\dots of a car wheel''), an exact-match
$3$-slot apple leaf, an attribute-drop $5$-slot pair of
green headphones (the color adjective ``green'' and the
disambiguating ``component'' are dropped from the predicted
caption), and a $6$-slot stringed-instrument base where
the prediction is more descriptive than the GT (``base
component'' $\to$ ``the body of a traditional Chinese
stringed instrument, likely a pipa or similar''), a case
in which Word-F1 penalizes plausibly added specificity.
All eight panels are from the held-out PartVerse-QA split
with object-disjoint training.

\subsection{3DCoMPaT-GrIn: full grounded descriptions}
\label{app:qual:grin}

\begin{figure}[!htbp]
  \centering
  \includegraphics[width=0.95\linewidth]{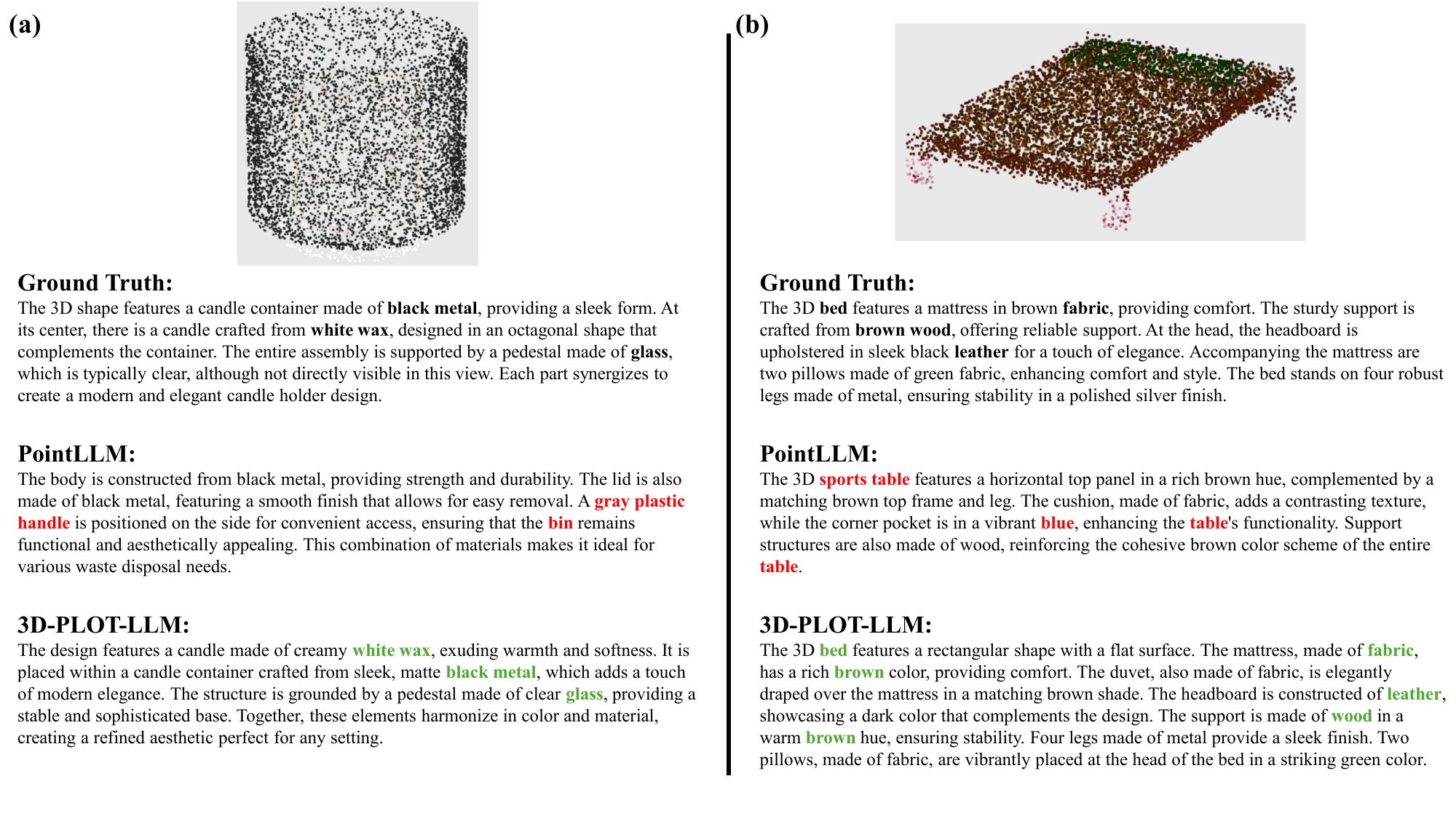}
  \caption{\textbf{3DCoMPaT-GrIn full grounded descriptions:
  Figures~\ref{fig:qualitative}(d-e) at full length.} Verbatim
  ground-truth, PointLLM, and 3D-PLOT-LLM descriptions for the candle
  holder (a) and bed (b) examples shown abridged in the main paper.
  PointLLM misclassifies both (bin, sports table) and the
  part-level materials/colors follow the misclassification
  (\textcolor{red}{red}); 3D-PLOT-LLM recovers the correct class and
  the multi-material attributes (\textcolor{green}{green}: matches GT).}
  \label{fig:appx-grin-set1}
\end{figure}

\begin{figure}[!htbp]
  \centering
  \includegraphics[width=0.95\linewidth]{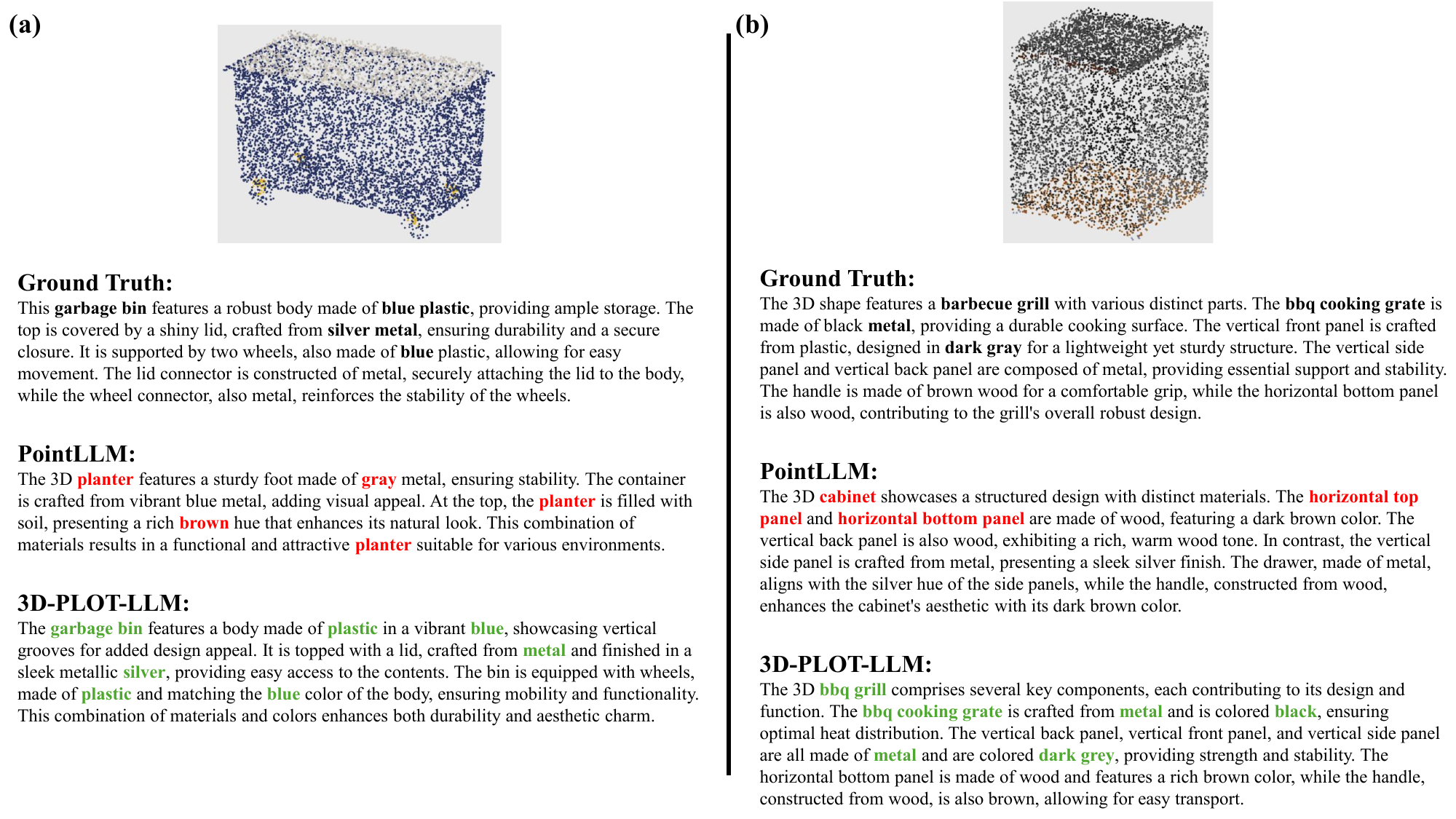}
  \caption{\textbf{3DCoMPaT-GrIn full grounded descriptions:
  Figure~\ref{fig:qualitative}(f) at full length plus an
  additional drift case.} (a) garbage bin (the third
  abridged main-paper panel) at full length, with PointLLM
  drifting to ``planter''. (b) a barbecue grill on which
  PointLLM drifts to ``cabinet'', confabulating horizontal top
  and bottom panels and the materials a cabinet would have.
  3D-PLOT-LLM stays on-class in both panels and recovers the
  multi-material attributes the ground truth describes.}
  \label{fig:appx-grin-set2}
\end{figure}

\begin{figure}[!htbp]
  \centering
  \includegraphics[width=0.95\linewidth]{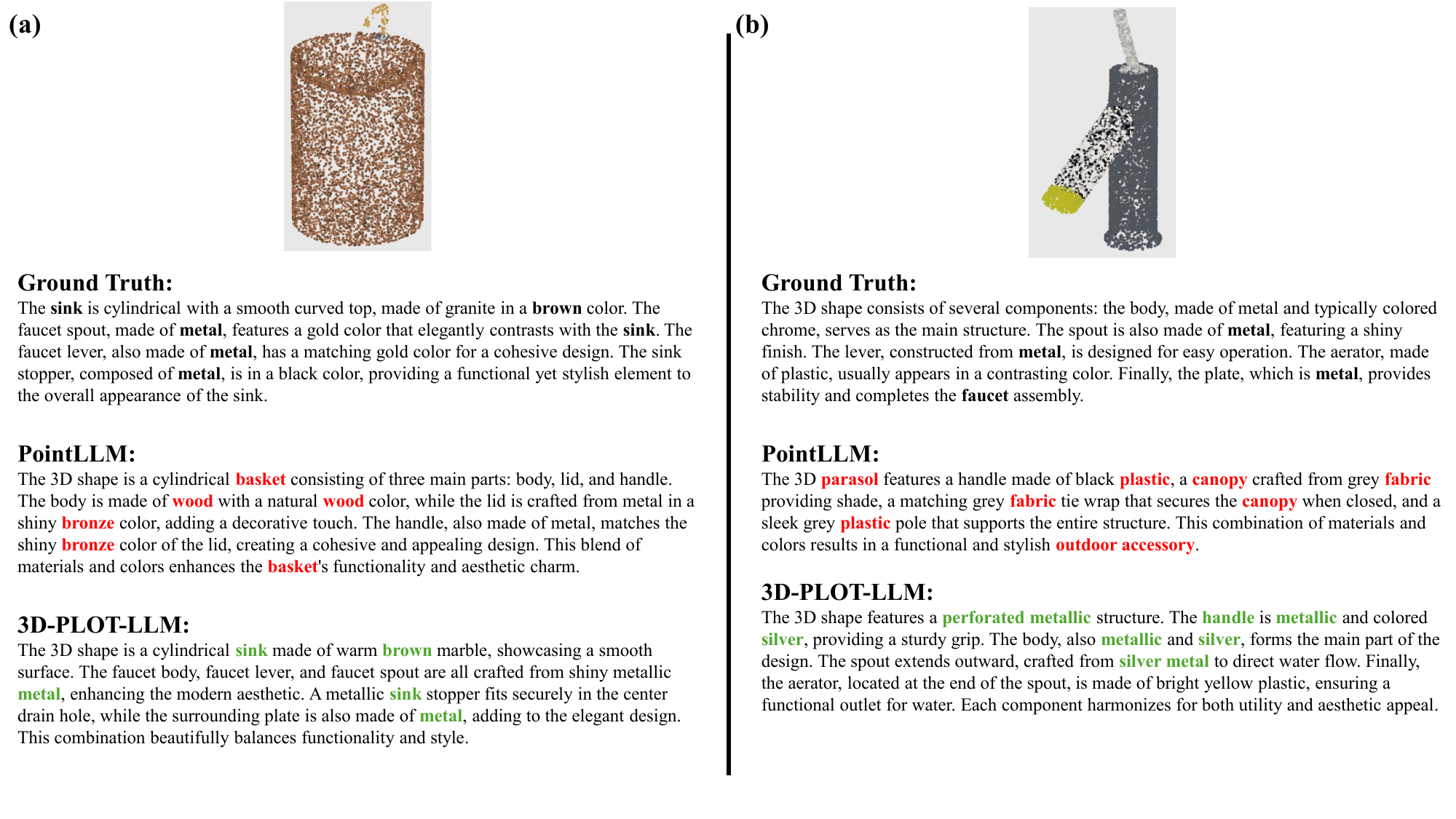}
  \caption{\textbf{3DCoMPaT-GrIn full grounded descriptions: two
  additional cases with maximally different drift targets.} (a) a
  cylindrical sink with attached metal faucet (spout, lever, stopper),
  misread by PointLLM as a wooden basket, a same-domain shape-driven
  drift; 3D-PLOT-LLM recovers the brown body and the metal
  faucet attachments. (b) a standalone metal faucet whose body,
  spout, and aerator are correctly described by 3D-PLOT-LLM
  in the metallic domain, while PointLLM hallucinates a parasol
  with fabric canopy and plastic pole. This faucet drift flips
  both material (metal $\to$ fabric) and use context
  (plumbing fixture $\to$ outdoor accessory).}
  \label{fig:appx-grin-set3}
\end{figure}

Figures~\ref{fig:appx-grin-set1}, \ref{fig:appx-grin-set2},
and~\ref{fig:appx-grin-set3} reproduce the three abridged
3DCoMPaT-GrIn panels of Figure~\ref{fig:qualitative}(d-f) at
full length and add three additional 3DCoMPaT-GrIn cases (a
barbecue grill, a sink, and a faucet) covering different drift targets. Each panel pairs the
input point cloud with the ground-truth, PointLLM, and
3D-PLOT-LLM grounded descriptions verbatim; green highlights
mark words that match the ground-truth class or
material/color attributes, and red highlights mark
hallucinated words that conflict with the ground truth. All
six cases share the same script: PointLLM, treating the
object as a flat token stream, drifts to a wrong object class
and confabulates parts that fit the misclassification (bin,
sports table, planter, cabinet, basket, parasol);
3D-PLOT-LLM, with structurally refined per-region markers,
recovers both the correct class and the material/color
attributes the ground truth describes.

\subsection{Objaverse: whole-object captioning side-by-side}
\label{app:qual:obja}

\begin{figure}[!htbp]
  \centering
  \includegraphics[width=0.95\linewidth]{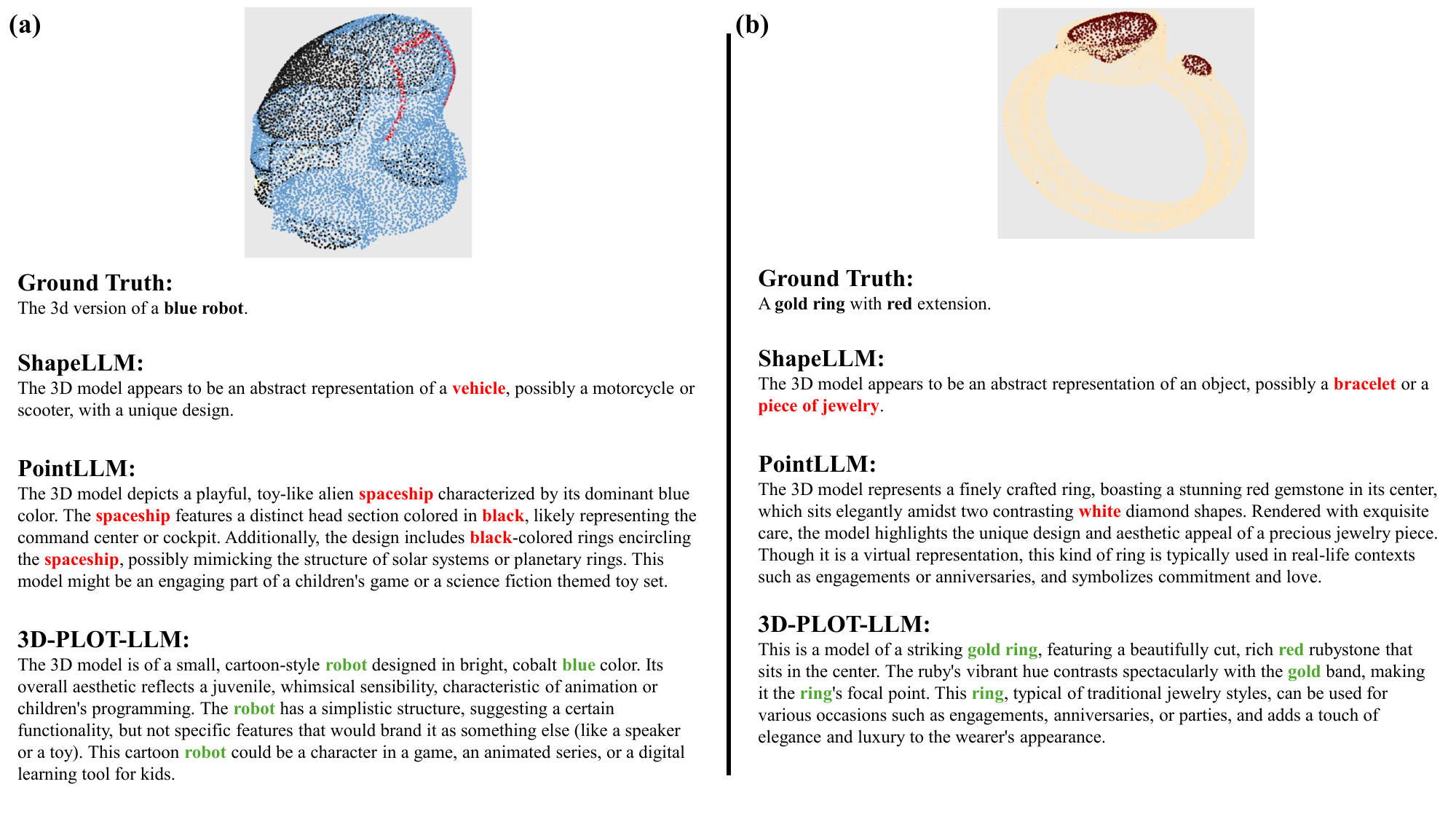}
  \caption{\textbf{Objaverse whole-object captioning: PointLLM
  failure modes.} Outputs from 3D-PLOT-LLM, PointLLM, and
  ShapeLLM on two held-out objects: (a) a blue robot
  mis-classed by PointLLM as ``spaceship'' (class drift); (b) a
  gold ring with a red gemstone, correctly classed by PointLLM
  but hallucinated as ``boasting two contrasting white diamond
  shapes'' that are absent from the point cloud (structural
  hallucination). 3D-PLOT-LLM stays anchored to both class and
  attributes in each case (\textcolor{green}{green}: matches GT;
  \textcolor{red}{red}: PointLLM hallucinations).}
  \label{fig:appx-obja}
\end{figure}

Figure~\ref{fig:appx-obja} shows the ground-truth,
3D-PLOT-LLM, PointLLM, and ShapeLLM outputs side by side,
surfacing two distinct PointLLM failure modes: class drift
(blue robot $\to$ ``spaceship'') and structural hallucination
(a gold ring described as ``boasting two contrasting white
diamond shapes'' that are absent from the point cloud), with
3D-PLOT-LLM staying anchored to both class and attributes in
each case. App.~\ref{app:qual:limits} item~(3) gives an
Objaverse failure mode where both models miss a $2$D-surface
attribute the point cloud does not carry.

\subsection{Failure modes}
\label{app:qual:limits}

\begin{figure}[!htbp]
  \centering
  \includegraphics[width=0.95\linewidth]{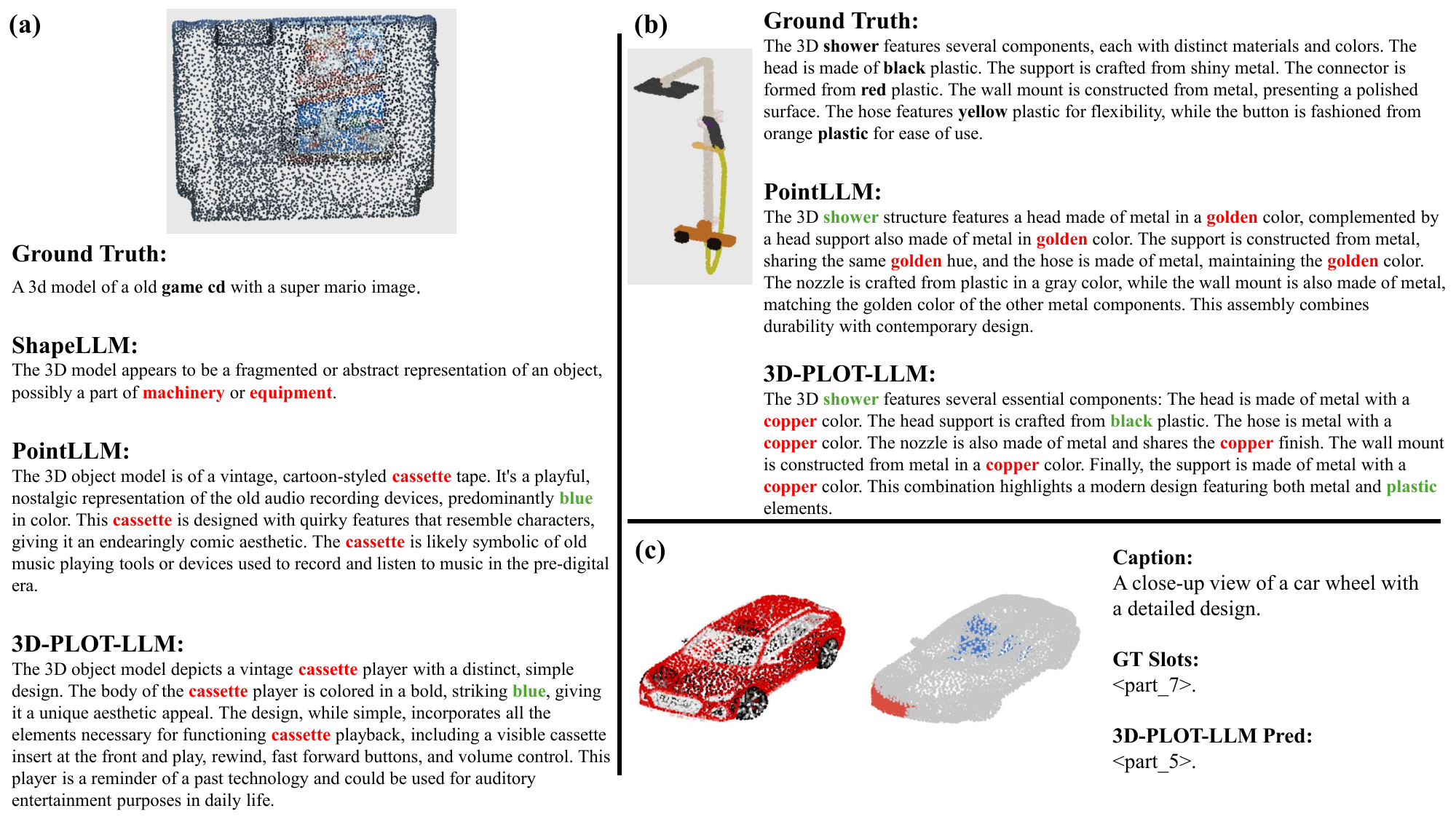}
  \caption{\textbf{Failure modes visualized across the three
  benchmarks.} (a) Mode~(3), Objaverse: both models call a
  Mario-image CD a \textcolor{red}{cassette}, recovering only the
  shared \textcolor{green}{blue} body color. (b) Mode~(2),
  3DCoMPaT-GrIn shower: both collapse the multi-color GT toward a
  single metal color (\textcolor{red}{golden} for PointLLM,
  \textcolor{red}{copper} for ours); 3D-PLOT-LLM still recovers
  \textcolor{green}{black plastic} on the head support that
  PointLLM omits. (c) Mode~(1), PartVerse-QA C2S: a caption
  ``\dots a car wheel with a detailed design'' has GT \partk{7}
  but the model commits to \partk{5}; the partition places
  different wheels in different slots.}
  \label{fig:appx-fail}
\end{figure}

We catalog three representative failure modes (visualized in
Fig.~\ref{fig:appx-fail}) and attribute each to its primary
cause: query-side language ambiguity, inherent benchmark
difficulty, and the frozen point encoder shared with the
PointLLM baseline.

\textbf{(1) Symmetric repeated parts.} On a PartVerse-QA C2S
car (Fig.~\ref{fig:appx-fail}c), the $K{=}16$ partition places
different wheels in different slots, but near-synonymous
captions targeting different wheels have no basis to pick one
over another and the model commits to a single slot regardless. The
reverse S2C direction (Fig.~\ref{fig:appx-partverse-s2c}, panel a)
recovers a clean caption when one wheel slot is supplied, isolating
the failure to query-side ambiguity rather than slot binding.

\textbf{(2) Both models degrade together.} On a 3DCoMPaT-GrIn
shower (Fig.~\ref{fig:appx-fail}b), both 3D-PLOT-LLM and
PointLLM collapse the GT's six distinct material/color
combinations toward a single metal color
(\textcolor{red}{copper} for ours, \textcolor{red}{golden} for
PointLLM). 3D-PLOT-LLM still recovers one non-metal element
(\textcolor{green}{black plastic} on the head support) that
PointLLM omits, but neither model recovers the full multi-color
spec. Two different architectures failing in the same direction
suggests the multi-color rendering is hard for the benchmark
itself, not a weakness specific to either model.

\textbf{(3) 2D-surface content not carried by the point cloud.}
Objaverse Mario-image CD (Fig.~\ref{fig:appx-fail}a): both models
see ``a blue rectangular vintage media object'' and converge on
\textcolor{red}{cassette}, missing the 2D Mario texture. The bound
is at the encoder level (frozen Point-BERT, shared with PointLLM)
and orthogonal to the part-vocabulary contribution.

\end{document}